\documentclass[11pt,a4paper]{article}

\usepackage{graphicx} 
\usepackage{pgf}
\usepackage{amssymb,amsmath,amsfonts}
\usepackage{bbm}
\usepackage{multirow}
\usepackage{booktabs}
\usepackage{color}
\usepackage[font=footnotesize]{caption}
\usepackage[font=footnotesize]{subcaption}
\usepackage{algorithm}
\usepackage[noend]{algpseudocode}
\usepackage{hyperref}
\usepackage{xcolor}
\definecolor{lnkcol}{rgb}{0,0,0.93}

\hypersetup{
    pdfauthor={Emmanuel Kalunga, Sylvain Chevallier and Quentin Barthelemy},     
    pdfsubject={Using Riemannian geometry for SSVEP-based Brain Computer Interface},   
    pdfnewwindow=true,      
    pdfkeywords={Riemannian geometry, Online, Asynchronous, Brain-Computer Interfaces, Steady State Visually Evoked Potentials}, 
    colorlinks=true,        
    linkcolor=lnkcol,       
    citecolor=lnkcol,       
}

\newcommand{\ra}[1]{\renewcommand{\arraystretch}{#1}}
\newcommand{\X}{\ensuremath{X}} 
\newcommand{\onlineX}{\ensuremath{\chi}} 
\newcommand{\Y}{\ensuremath{Y}} 
\newcommand{\totd}{\ensuremath{{m}}} 
\newcommand{\chI}{\ensuremath{c}} 
\newcommand{\dc}{\ensuremath{C}} 
\newcommand{\si}{\ensuremath{n}} 
\newcommand{\dt}{\ensuremath{N}} 
\newcommand{\dr}{\ensuremath{\mathcal{N}}} 
\newcommand{\nb}{\ensuremath{i}} 
\newcommand{\Nb}{\ensuremath{I}} 
\newcommand{\ci}{\ensuremath{k}} 
\newcommand{\dK}{\ensuremath{K}} 
\newcommand{\df}{\ensuremath{f}} 
\newcommand{\dF}{\ensuremath{F}} 
\newcommand{\ti}{\ensuremath{i}} 
\newcommand{\dT}{\ensuremath{I}} 
\newcommand{\setindex}{\ensuremath{\mathcal{I}}} 
\newcommand{\dd}{\ensuremath{D}} 
\newcommand{\tj}{\ensuremath{j}} 
\newcommand{\Dindex}{\ensuremath{\mathcal{J}}} 
\newcommand{\cue}{\ensuremath{\tau_0}} 
\newcommand{\deltaN}{\ensuremath{\Delta \si}} 
\newcommand{\ws}{\ensuremath{w}} 
\newcommand{\di}{\ensuremath{d}} 
\newcommand{\Xs}[1] {\ensuremath{\X_{\operatorname{\text{#1}}}}} 

\newcommand{\GenMa}{\ensuremath{\mathcal{M}}} 
\newcommand{\Ma}{\ensuremath{\mathcal{M}_{\dc}}} 
\newcommand{\Sy}{\ensuremath{\mathcal{S}_{\dc}}} 
\renewcommand{\P}{\ensuremath{\Sigma}} 
\renewcommand{\S}{\ensuremath{\Theta}} 
\newcommand{\ev}{\ensuremath{\lambda}} 
\newcommand{\distR}{\ensuremath{\delta}} 
\newcommand{\Rm}{\ensuremath{\mu}} 
\newcommand{\EX}{\ensuremath{\omega}} 
\newcommand{\tarco}{\ensuremath{\Gamma}} 
\newcommand{\mle}{\ensuremath{\hat{\ell}}} 
\newcommand{\fmle}{\ensuremath{g}} 
\newcommand{\pocc}{\ensuremath{\rho}} 
\newcommand{\pthres}{\ensuremath{\vartheta}} 

\newcommand{\clout}{\ensuremath{\ci^*}} 
\newcommand{\cloutOn}{\ensuremath{\tilde{\ci}}} 
\newcommand{\cloutmp}{\ensuremath{\bar{\ci}}} 
\newcommand{\clarray}{\ensuremath{\mathcal{K}}} 

\newcommand{\x}{\ensuremath{x}} 
\newcommand{\xb}{\ensuremath{\bar{x}}}
\newcommand{\eye}{\ensuremath{\mathbbm{I}}}
\newcommand{\unitary}{\ensuremath{\mathbbm{1}}}
\newcommand{\argmin}{\arg\!\min}
\DeclareMathOperator*{\argmax}{arg\,max}
\renewcommand{\Re}{\ensuremath{\mathbb{R}}}
\newcommand{\Log}[1]{\ensuremath{\operatorname{Log}_{#1}}}
\newcommand{\Exp}[1]{\ensuremath{\operatorname{Exp}_{#1}}}
\newcommand{\Expm}{\ensuremath{\operatorname{Exp}}}
\newcommand{\Logm}{\ensuremath{\operatorname{Log}}}
\newcommand{\cov}[1]{\ensuremath{\hat{\Sigma}_{\operatorname{#1}}}}
\newcommand{\disvec}{\ensuremath{\widetilde{\delta}}} 
\newcommand{\diffdist}{\ensuremath{\delta}}  

\begin{document}

\begin{titlepage}

\newcommand{\HRule}{\rule{\linewidth}{0.5mm}} 

\center 
 

\textsc{\LARGE Tshwane University of Technology}\\ 
\textsc{\LARGE \&}\\ 
\textsc{\LARGE Universit\'e de Versailles Saint-Quentin}\\ 
\textsc{\LARGE \&}\\ 
\textsc{\LARGE Mensia Technologies}\\[1.5cm] 


\HRule \\[0.4cm]
{ \huge \bfseries Research Report:\\Using Riemannian geometry for SSVEP-based Brain Computer Interface}\\[0.4cm] 
\HRule \\[1.5cm]
 

\begin{minipage}{0.4\textwidth}
\begin{flushleft} \large
\emph{Authors:}\\
Emmanuel \textsc{Kalunga}\\[1.3cm] 
Sylvain \textsc{Chevallier}\\[1.5cm]
Quentin \textsc{Barth\'elemy}\\
\end{flushleft}
\end{minipage}
~
\begin{minipage}{0.4\textwidth}
\begin{flushright} \small
\emph{Affiliations:} \\
Department of Electrical Engineering, F'SATIE,  TUT, Pretoria, South Africa\\[0.5cm]
Laboratoire d'Ing\'{e}nierie des Syst\`{e}mes de Versailles, UVSQ, Velizy, France\\[0.5cm]
Mensia Technologies, ICM, H\^{o}pital de la Piti\'{e}-Salp\^{e}tri\`{e}re, Paris, France\\
\end{flushright}
\end{minipage}\\[1cm]



{\large \today}\\[1cm] 


\pgfimage[interpolate=true,width=0.15\linewidth]{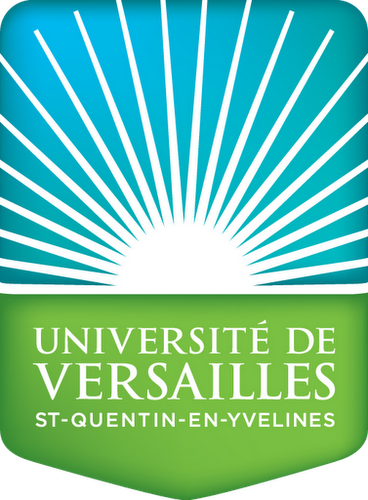} 
\hfill
\pgfimage[interpolate=true,width=0.2\linewidth]{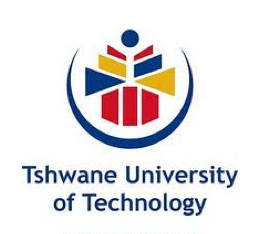} 
\hfill
\pgfimage[interpolate=true,width=0.2\linewidth]{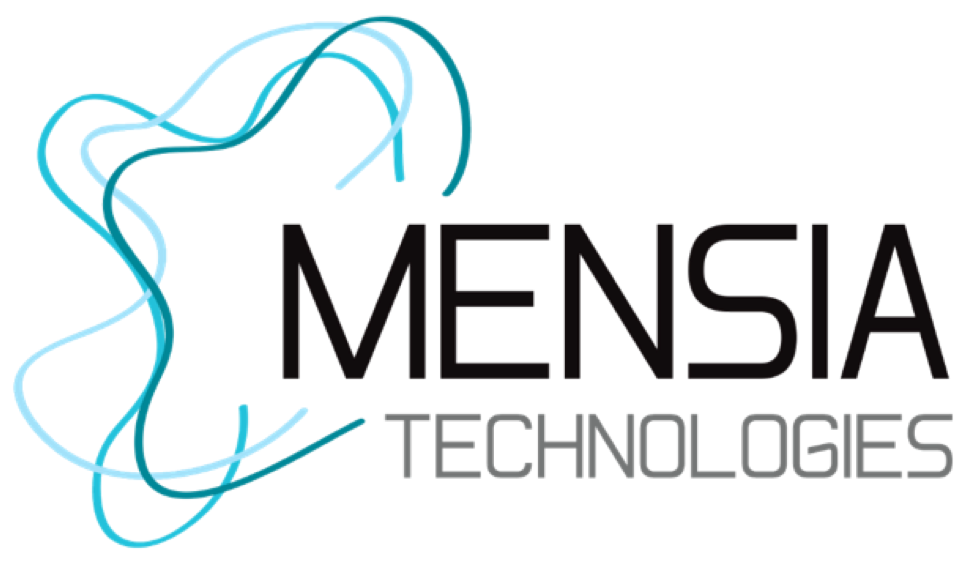} 
 

\vfill 

\end{titlepage}

\title{Using Riemannian geometry for SSVEP-based Brain Computer Interface}
\author{
  Emmanuel K. Kalunga\thanks{Department of Electrical Engineering, French South African Institute of Technology, Tshwane University of Technology, Pretoria 0001, South Africa. \texttt{emmanuelkalunga.k@gmail.com}} \and
  Sylvain Chevallier\thanks{Laboratoire d'Ing\'{e}nierie des Syst\`{e}mes de Versailles, Universit\'{e} de Versailles Saint-Quentin, Velizy 78140, France. \texttt{sylvain.chevallier@uvsq.fr}} \and
  Quentin Barth\'{e}lemy\thanks{Mensia Technologies, ICM, H\^{o}pital de la Piti\'{e}-Salp\^{e}tri\`{e}re, 75013 Paris, France. \texttt{quentin.barthelemy@mensiatech.com}}
}

\maketitle

\begin{abstract}
Riemannian geometry has been applied to Brain Computer Interface (BCI) for brain signals classification yielding promising results. 
Studying electroencephalographic (EEG) signals from their associated covariance matrices allows a mitigation of common sources of variability (electronic, electrical, biological) by constructing a representation which is invariant to these perturbations. 
While working in Euclidean space with covariance matrices is known to be error-prone, one might take advantage of algorithmic advances in information geometry and matrix manifold to implement methods for Symmetric Positive-Definite (SPD) matrices. 
This paper proposes a comprehensive review of the actual tools of information geometry and how they could be applied on covariance matrices of EEG.
In practice, covariance matrices should be estimated, thus a thorough study of all estimators is conducted on real EEG dataset.
As a main contribution, this paper proposes an online implementation of a classifier in the Riemannian space and its subsequent assessment in Steady-State Visually Evoked Potential (SSVEP) experimentations.
\end{abstract}


\section{Introduction}

Human-machine interactions without relying on muscular capabilities is possible with Brain-Computer Interfaces (BCI)~\cite{wolpaw2002bcireview,PFU10,ZAN11}.
They are the focus of a large scientific interest~\cite{VID73,BAY00,LOP09,HUA11,JRA11,TU12}, especially those based on electro-encephalography (EEG) \cite{NIE04,Blankertz2006}.
From the large literature based on the BCI competition datasets~\cite{BLA04,BLA06,TAN12}, one can identified the two most challenging BCI problems: on one hand, the inter-individual variability plagues the models and lead to BCI-inefficiency effect~\cite{ALL10,VID10,HAM12}, on the other hand the intra-individual changes calls for the development of online algorithms and adaptive systems following the evolution of the subject's brain waves~\cite{Obermaier2001,SHE06,Lenhardt2008,Bin2009AnOnlineMultiCh,ZHU11}. 
To alleviate these variations, several signal processing and machine learning techniques have been proposed, such as filtering, regularization or clustering~\cite{Lee2003,kalunga2013ssvep,Lu2010Regular} without marking out an obvious ``best candidate'' methodology. 

A common vision is shared by all the most efficient approaches to reduce signal variabilities: they are applied on covariance matrices instead of working in the input signal space.
Common Spatial Pattern (CSP)~\cite{BLA08b,LOT11,ANG12,YAN12a}, which is the most known preprocessing technique in BCI, try to maximize the covariance of one class while minimizing the covariance of the other one.
Similarly, Principal Components Analysis (PCA)~\cite{BLA04,BLA06}, also applied for spatial filtering in BCI, is based on the estimation of covariance matrices.
Canonical Correlation Analysis (CCA)~\cite{Hardoon2004} is another example of a technique relying on covariance estimates successfully applied on EEG for spatial filtering~\cite{Lin2007,Bin2009AnOnlineMultiCh,kalunga2013ssvep}. 
Given two sets of signals, CCA aims at finding the projection space that maximizes their cross-covariance while jointly minimizing their covariance.
Covariance matrices are also found in classifiers such as the Linear Discriminant Analysis (LDA), which is largely used in BCI. 
In all cases, they are handled as elements of an Euclidean space.
However, being Symmetric and Positive-Definite (SPD), covariance matrices lie on a subset of the Euclidean space, with reduced dimensionality and specific properties, the \emph{Riemannian manifold}. 
Considering covariance matrices in their original space would reduce the search area for an optimization problem \cite{absil2009optimization}. 
As Riemannian manifolds inherently define a metric, the distance between SPD matrices takes into account the space where they lie on; approximating it to an Euclidean space introduce inaccuracies and result in ill-conditioned matrices.  

Recently, studies have been done to consider covariance matrices obtained from multichannel brain signals in their original space \cite{barachant2010riemannian, barachant2012multiclass}. 
Covariance matrices are the input features of the BCI system and the classifier algorithms rely on Riemannian metric for partitioning the feature space.
The authors propose to build specific covariance matrices in order to emphasize the spatial and temporal information of the multichannel brain signals.
The outcome of this approach is a simple processing toolchain, which achieve state-of-the-art classification performances.

This paper introduces an online version of the \emph{minimum distance to Riemannian mean} algorithm~\cite{barachant2012multiclass}, with an application on Steady-State Visually Evoked Potential (SSVEP) recordings. 
In SSVEP, the subjects concentrate on stimuli blinking with fixed frequencies; depending on the focus of their attention, brain waves will arise with the same phase and frequency than the stimulus chosen by the subject.
The proposed online implementation improves the performances in the BCI task, as the parameters of the classifier adapt to new, unlabeled samples.
Such adaptation takes into account the qualitative changes in the brain responses and the acquired signals, caused respectively by internal and external factors, e.g. user fatigue and slight variations in experimental settings.
The proposed online implementation is similar to the unsupervised or semi-unsupervised learning scheme proposed in~\cite{panicker2010adaptation,schettini2014self}; that has the potential of shortening (or even removing) the calibration phase. 
The proposed algorithm applies a similar approach to dynamic stopping criterion used in \cite{verschore2012dynamic} to increase the speed of the BCI system.
This approach allows to dynamically determine the trial length and ensure robustness in classification results.    
When working with covariance matrices, a crucial point is to correctly estimate the covariance when the number of samples is small or heavily corrupted by noise. 
Several approaches have been proposed to build the covariance matrices, relying on normalization or regularization of the sample covariances. 
To assess the quality of the covariance matrices obtained from EEG samples, a comparative study of these estimators is conducted. 

Hence, the contributions of this works are:
\begin{itemize}
\item a comprehensive review of the literature on Riemannian geometry applied to EEG and time-series,
\item an online and asynchronous classification algorithm for SSVEP-based BCI,
\item a thorough analysis of the covariance estimators and their impact on tools derived from information geometry.
\end{itemize}

The paper is divided as follows: 
Section~\ref{sec:stateoftheart} reviews application of Riemannian geometry to machine learning. 
Section~\ref{sec:riem_geom} presents concepts of Riemannian geometry relevant to this work and estimators of covariance.
In Section~\ref{sec:proposed_algo}, the proposed classification algorithm for online SSVEP is introduced and the experimental results are presented in Section~\ref{sec:exp_valid}.    


\section{State of the Art}
\label{sec:stateoftheart}

Information geometry provides useful tools for various machine learning and optimization problems. 
In machine learning, SPD matrices 
 have been used in various applications where features and data  are only considered in the Euclidean space. 
Indeed, covariance matrices lie in the space of SPD matrices which is a subset of the Euclidean space when considered with the scalar product.
But the same space of SPD matrices, endowed with a differential structure, induces a Riemannian manifold.



Riemannian geometry can improve machine learning algorithms, taking explicitly into consideration the underlying structure of the considered space.
Three kind of approaches in the literature uses the data geometry in machine learning. 
The first one relies on the mapping of the Riemannian manifold onto an Euclidean vector space. 
One such mapping, called logarithmic mapping, exist between the manifold and its tangent space, which is an Euclidean space, and has been used in classification task for BCI~\cite{barachant2012bci,BAR13}. 
Other kernels have been applied successfully to this end: Stein kernel, Log-Euclidean kernels as well as their normalized versions~\cite{yger2013review}.
The main idea is to map the input data to a high dimensional feature space, providing a rich and hopefully linearly separable representation.
The so-called kernel trick is to provide a kernel function, which computes an inner product in the feature space directly from points lying in the input space, defining a Reproducing Kernel Hilbert Space (RKHS).
The family of kernels defined on the Riemannian manifold allows implementing extension of all kernel-based methods, such as SVM, kernel-PCA or kernel $k$-means~\cite{jayasumana2013kernel}.
Apart from the kernel approaches, once the data are mapped onto a vector space, any machine learning algorithm working in Euclidean space, such as LDA, could be applied~\cite{barachant2012multiclass}.

A second kind of machine learning approach exploit the underlying geometry of the data.
Instead of mapping the data to an Euclidean space, either a tangent space or a RKHS, the algorithms are adapted to Riemannian space. 
For instance, sparse coding algorithm have been adapted to Riemannian manifold, using the geodesic distance to estimate the data point and its sparse estimate~\cite{xie2013nonlinear}.
Similarly nonlinear dimensionality reduction techniques have been adapted to Riemannian manifold, such as Laplacian Eigenmaps (LE), Locally Linear Embedding (LLE), and Hessian LLE. 
This adaptation was used to cluster data using their pdfs \cite{goh2008unsupervised} or covariance matrices \cite{goh2008clustering} as features. 
Another example is the adaptation of interpolation and filtering of data to Riemannian space performed in \cite{PEN06}, where an affine-invariant Riemannian metric is also proposed to offer a geodesically complete manifold i.e a manifold with no edge and no singular point that can be reached in finite time.  

In the last kind of approach, instead of adapting existing algorithm from Euclidean to Riemannian geometry, new algorithms are developed directly for Riemannian manifolds.
The \emph{minimum distance to Riemannian mean} (MDRM) relies on a Riemannian metric to implement a multi-class classifier and have been applied on EEG.
New EEG trials are assigned to the class whose average covariance matrix is the closest to the trial covariance matrix~\cite{barachant2012multiclass}.
The MDRM classification can be preceded by a filtering of covariance matrices, like in~\cite{barachant2010riemannian} where covariance matrices are filtered with LDA component in the tangent space, then brought back to the Riemannian space for classification with MDRM. 
Another example is the \emph{Riemannian Potato} \cite{barachant2013riemannian}, an unsupervised and adaptive artifact detection method, providing an online adaptive EEG filtering (i.e outliers removal). 
Incoming signals are rejected if their covariance matrix lies beyond a predefined distance z-score from the mean covariance matrix, computed from a sliding window.
With the same objective of achieving robustness to noise that affect covariance matrices, Riemannian geometry is used to solve divergence functions of pdfs~\cite{amari2010information}.
This allows to reformulate the CSP as the maximization of the divergence between the distributions of data from two different classes corresponding to two cognitive states~\cite{samek2013robust, samek2014information}. 
Using the \emph{beta divergence} the obtained CSP is robust to outliers in sample covariance matrices and this algorithm is successfully applied to EEG filtering for BCI.   
Riemannian metrics are also used for EEG channel selection~\cite{barachant2011channel} and the selection of the most discriminatory spatial filters in CSP~\cite{barachant2010common}.  



Applications of Riemannian geometry to BCI mentioned thus far are focusing on motor imagery (MI) paradigm.
In MI experiment, the subject is asked to imagine a movement (usually hand, feet or tongue), generating Event-Related Synchronization and Desynchronization (ERD/ERS) in pre-motor brain area.
Riemannian BCI is well suited for MI experiment as the spatial information linked with synchronization is directly embedded in covariance matrices obtained from multichannel recordings.
However, for BCI that rely on Evoked Potential such as SSVEP or Event Related Potential (ERP), as P300, both frequential and temporal information are needed; the spatial covariance matrix does not contain these information. 
To apply Riemannian geometry to SSVEP and ERP, the sample covariance matrices can be defined from a rearrangement of the recorded data. 
The rearrangement is done such that the temporal or frequency information are captured~\cite{congedo2013new}. 
With similar motivations, \cite{li2009eeg} and \cite{li2012electroencephalogram} defined a new Riemannian distance between SPD matrices that would take into account a weighting factor on matrices. 
They use this new distance as a dissimilarity between weighted matrices of power spectral density to classify EEG into different sleep state by $k$-nearest neighbors. 


\section{Covariance Matrices and their Geometry}
\label{sec:riem_geom}

This section presents some formal definitions for the information geometry concepts used in this paper.
The link with the covariance matrices is explicated in Section~\ref{sec:cov_mat}, along with the covariance estimators proposed in the literature. 

\subsection{Riemannian Manifold}
\label{sec:notdef}

An $\totd$-dimensional \emph{manifold} $\GenMa$ is a Hausdorff space for which every point has a neighborhood that is homeomorphic to an open subset of $\Re^\totd$ \cite{lee2010introduction, jost2011riemannian}. 
When a tangent space is defined at each point, $\GenMa$ is called a differential manifold. 
A \emph{geodesic} $\gamma$ is the shortest smooth curve between two points, $\P_1$ and $\P_2$. 
The tangent space $T_\P\GenMa$ at point $\P$ is the vector space spanned by the tangent vectors of all geodesics on $\GenMa$ passing through $\P$. 
A \emph{Riemannian} manifold is a manifold endowed with an inner product defined on the tangent space, which varies smoothly from point to point.  

For the rest of this paper, we will restrict to the analysis of the manifold $\Ma$ of  the $\dc\times\dc$ symmetric positive definite matrices, defined as:
\begin{equation}
	\label{eq:rm}
  \Ma = \left\{ \P \in  \Re^{\dc\times\dc} : \P = \P^\intercal \text{ and } x^\intercal \P x > 0, \forall x \in \Re^{\dc} \backslash 0 \right\} \ . \nonumber
\end{equation}
The tangent space $T_{\P}\Ma$ is identified to the Euclidean space of symmetric matrices:
\begin{equation}
  \label{eq:sy}
  \Sy = \left\{ \S \in \Re^{\dc\times\dc} : \S = \S^\intercal \right\} \ . \nonumber
\end{equation}
The dimension of the manifold $\Ma$, and its tangent space $T_{\P}\Ma$, is $\totd = \dc(\dc+1)/2$.

The mapping from a point $\S_i$ of the tangent space to the manifold is called the exponential mapping $\Exp{\P}(\S_i)$: $T_{\P}\Ma \rightarrow \Ma$ and is defined as:
\begin{equation}
	\label{eq:exp_r}
	\Exp{\P}(\S_i) = \P^{\frac{1}{2}}\Expm(\P^{-\frac{1}{2}}\S_i \P^{-\frac{1}{2}})\P^{\frac{1}{2}} \ .
\end{equation}  
Its inverse mapping, from the manifold to the tangent space is the logarithmic mapping $\Log{\P}(\P_i)$: $\Ma \rightarrow T_{\P}\Ma$ and is defined as:
\begin{equation}
	\label{eq:log_r}
	\Log{\P}(\P_i) = \P^{\frac{1}{2}}\Logm(\P^{-\frac{1}{2}}\P_i \P^{-\frac{1}{2}})\P^{\frac{1}{2}} \ .
\end{equation} 
$\Expm(\cdot)$ and $\Logm(\cdot)$ are the the matrix exponential and logarithm respectively.
The computation of these operators is straightforward for SPD matrices of $\Ma$. 
They are obtained from their eigenvalue decomposition (EVD): 
\begin{displaymath}
  \begin{split}
    \P &= U\, \text{diag}(\ev_1, \dots, \ev_\dc)\, U^\intercal \ , \\
    \Expm(\P) &= U\, \text{diag}(\log(\ev_1), \dots, \log(\ev_\dc))\, U^\intercal \ , \\
    \Logm(\P) &= U\, \text{diag}(\exp(\ev_1), \dots, \exp(\ev_\dc))\, U^\intercal \ ,
  \end{split}
\end{displaymath}
where $\ev_1, \dots, \ev_\dc$ are the eigenvalues and $U$ the matrix of eigenvectors of $\P$.
As any SPD matrix can be diagonalized with strictly positive eigenvalues, $\Logm(\cdot)$ is always defined.
Similarly the square root $\P^{\frac{1}{2}}$ is obtained as:
\begin{equation}
  \label{eq:sqm}
  \P^{\frac{1}{2}} = U\, \text{diag}(\ev_1^{\frac{1}{2}}, \dots, \ev_\dc^{\frac{1}{2}})\, U^\intercal \ , \nonumber
\end{equation}
and is unique. The same goes for $\P^{-\frac{1}{2}}$.

The tangent vector of the geodesic $\gamma(t)$ between $\P_1$ and $\P_2$, where $\gamma(0) = \P_1$ and $\gamma(1) = \P_2$ is defined as $v = \overrightarrow{\P_1 \P_2} = \Log{\P_1}(\P_2)$. 
A Riemannian distance between $\P_1$ and $\P_2$ can thus be defined as~\cite{moakher2005differential}:  
\begin{equation}
	\label{eq:dist_r}
	\distR(\P_1,\P_2) = \|\Logm(\P_1^{-1}\P_2)\|_F = \left[\sum_{\chI=1}^{\dc}\log^{2}\lambda_{\chI}\right]^{1/2} ,
\end{equation}
where $\lambda_\chI$, $\chI = 1, \dots, \dc$, are the eigenvalues of $\P_1^{-1}\P_2$.
From Eq.~\eqref{eq:dist_r}, the geometric mean of $\Nb$ points $\P_\nb$ on the manifold, $\nb = 1, \dots , \Nb$, can be defined as the point that minimizes the sum of squared distances to all $\P_\nb$:
\begin{equation}
\label{mean_r}
\Rm(\P_1, \dots, \P_\Nb) = \argmin_{\P \in \Ma} \sum_{\nb=1}^{\Nb} \distR^2(\P_\nb,\P) \ .
\end{equation}
This geometric mean has no closed form, and can be computed iteratively \cite{fletcher2004principal}.

\subsection{Covariance Matrix Estimation}
\label{sec:cov_mat}

Let $\x_\si \in \Re^{\dc}$, $\si=1, \ldots, \dt$, denotes a sample of a multichannel EEG trial recorded on $\dc$ electrodes. 
$\dt$ is the trial length. 
Let $\X \in \Re^{\dc\times\dt}$ be the EEG trial such as $\X = [\x_1, \ldots, \x_{\dt} ]$.
Under the hypothesis that all $\dt$ samples $\x_\si$ are randomly drawn from a distribution, it follows that $\X$ is a variable of random vectors and its expected vector is $\EX = E\{ \X \}$ \cite{fukunaga1990introduction}. 
The covariance matrix of the random vector $\X$ is defined by $\P = E\{ (\X - \EX)(\X - \EX)^{\intercal} \}$ and is unknown, thus an estimate $\cov{ }$ should be computed.
The choice of the appropriate estimator is crucial to verify that the obtained covariance matrices fulfill the following properties: they should be accurate, SPD and well-conditioned.
The last property requires that the ratio between the maximum and minimum singular value is not too large.
Moreover, to ensure the computational stability of the algorithm, the estimator should provide full-rank matrices, and its inversion should not amplify estimation errors. 

\subsubsection{Sample Covariance Matrix Estimator}

The most usual estimator is the empirical \emph{sample covariance matrix} (SCM), defined as: 
\begin{equation}
  \begin{split}
    \cov{scm} & = \frac{1}{\dt-1} \sum_{\si=1}^{\dt} (\x_\si - \xb)(\x_\si - \xb)^\intercal 
		\\
    & = \frac{1}{\dt-1} \X\left( \eye_\dt -\frac{1}{\dt}\unitary_{\dt} \unitary_{\dt}^\intercal \right) \X^\intercal \ ,
  \end{split}  
\end{equation}
where $\xb$ is the sample mean vector $\xb = \frac{1}{\dt} \sum_{n=1}^\dt \x_\si $.
In the matrix notation, $\eye_{\dt}$ is the $\dt\times\dt$ identity matrix and $\unitary_{\dt}$ is the vector $[1, \ldots, 1]$.
The SCM is often normalized as:
\begin{equation}
  \label{eq:nscme}
  \cov{nscm} = \frac{\dc}{\dt} \sum_{\si=1}^{\dt} \frac{(\x_\si - \xb)(\x_\si - \xb)^\intercal }{\sigma_{\x_\si}^2} \ ,
\end{equation}
$\sigma_{\x_\si}^2$ is the inter-channel variance at time $\si$. Other normalization techniques could be used. 

This estimation is fast and computationally simple.
However when $\dc \approx \dt$, the SCM is not a good estimator of the true covariance. 
In the case $\dc > \dt$, the SCM is not even full rank.

\subsubsection{Shrinkage Covariance Matrix Estimators}

To overcome these shortcomings, the shrinkage estimators have been developed as a weighted combination of the SCM and a target covariance matrix, which is often chosen to be close to the identity matrix, \textit{i.e.} resulting from almost independent variables of unit variance.
\begin{equation}
	\label{eq:shrink}
	\cov{shrink} = \kappa \tarco +(1-\kappa)\cov{scm} \ ,
\end{equation} 
where $0 \leqslant \kappa < 1$.
This estimator provides a regularized covariance that outperforms the empirical $\cov{scm}$ for small sample size, that is $\dc \approx \dt$. 
The shrinkage estimator has the same eigenvectors as the SCM, but the extreme eigenvalues are modified \textit{i.e.} the estimator is shrunk or elongated toward the average.

The different shrinkage estimators differ in their definition of the target covariance matrix $\tarco$. 
Ledoit and Wolf~\cite{ledoit2004well} have proposed $\tarco=\mathit{v}\eye_{\dc}$, with $\mathit{v} = \text{Tr}(\cov{scm} \eye_{\dc})$. 
Blankertz~\cite{blankertz2011single} defines $\tarco$ also as $\mathit{v}\eye_{\dc}$ but with $\mathit{v} = \frac{\text{Tr}(\cov{scm})}{\totd}$, $\totd$ being the dimension of the covariance matrices space. 
Sch\"{a}fer proposes several ways of defining $\tarco$ depending on the observed $\cov{scm}$~\cite{schafer2005shrinkage}.

%

\subsubsection{Fixed Point Covariance Matrix Estimator}

The Fixed Point Covariance Matrix~\cite{pascal123theoretical} is based on the maximum likelihood estimator $\mle$ which is a solution to the following equation:
\begin{equation}
\label{eq:max_lik}
\cov{fp} = \mle = \frac{\dc}{\dt} \sum_{\si=1}^\dt \left( \frac{(\x_\si - \xb)(\x_\si - \xb)^\intercal }{(\x_\si - \xb)^\intercal  \mle^{-1} (\x_\si - \xb)} \right) \ .
\end{equation}
As there is no closed form expression to Eq.~\eqref{eq:max_lik}, it can be written as a function of $\mle$: $\fmle(\mle) = \cov{fp}$.
$\fmle$ admits a single \emph{fixed point} $\mle^*$, where $\fmle(\mle^*) = \mle^*$, which is a solution to Eq.~\eqref{eq:max_lik}. 
Using $\mle_{0}:=\cov{nscm}$ as the initial value of $\mle $, it is solved recursively as $\mle_{t} \underset{t \to \infty}{\longrightarrow} \mle^{*}$.


%
%




\section{Online Adaptation\\of the Riemannian Classifier}
\label{sec:proposed_algo}

Concerning Riemannian classification of SSVEP, the offline methodology has been explained in \cite{congedo2013new}.
In this paper, we propose an online classifier for SSVEP, composed of an offline training step and an online and asynchronous test step.
This analysis is performed for each subject independently.

\subsection{Offline Riemannian Classification}

The proposed classifier relies on the Minimum Distance to Riemannian Mean (MDRM) introduced in~\cite{barachant2010riemannian} and extended in~\cite{congedo2013new} for possible applications on SSVEP signals.
Let consider an experimental SSVEP setup with $\dF$ stimulus blinking at $\dF$ different frequencies. It is a multiclass classification with $\dK=\dF+1$ classes: one class per stimulus and one resting state class.
The covariance matrices are estimated from a modified version of the input signal $\X$: 
\begin{equation}
\X \in \Re^{\dc \times \dt} \rightarrow 	
\begin{bmatrix}
 X_{\text{freq}_1}\\ 
\vdots \\
X_{\text{freq}_{\dF}} \\
\end{bmatrix}
\in \Re^{\dF\dc \times \dt} \ ,
\label{eq:ext_data}
\end{equation}
where $X_{\text{freq}_\df}$ is the input signal $\X$ band-pass filtered around frequency $\text{freq}_\df$, $\df=1, \ldots, \dF$. 
Thus the resulting covariance matrix $\cov{}$ belongs to $\Re^{\dF\dc \times \dF\dc}$. Henceforth, all EEG signals will be considered as filtered and modified by Eq.~\eqref{eq:ext_data}.

From $\dT$ labelled training trials $ \left\{ \X_{\ti} \right\}_{\ti=1}^{\dT}$ recorded per subject, $\dK$ centers of classes $\P_{\Rm}^{(\ci)}$ are estimated using Algorithm~\ref{alg:r_mean}. 
When an unlabelled test trial $\Y$ is given, it is classified as belonging to the class whose center $\P_{\Rm}^{(\ci)}$ is the closest to the trial's covariance matrix (Algorithm~\ref{alg:mdrm}, step \ref{op:decision}).

\begin{algorithm}
\caption{Offline Estimation of Riemannian Centers of Classes}
\label{alg:r_mean}
	Inputs: $\X_{\ti} \in \Re^{\dF \dc\times\dt}$, for $\ti = 1, \ldots, \dT$, a set of labelled trials. \\
	Inputs: $\setindex(\ci)$, a set of indices of trials belonging to class $\ci$. \\
	Output: $\P_{\Rm}^{(\ci)}$, $\ci = 1, \ldots, \dK$, centers of classes.
	\begin{algorithmic}[1]
	\State Compute covariance matrices $\cov{\textit{\ti}}$ of $\X_\ti$ 
	\label{op:cov_i}
	\State \textbf{for} $\ci$ = 1 \textbf{to} $\dK$ \textbf{do}
	\State \quad $\P_{\Rm}^{(\ci)}=\Rm(\cov{\textit{\ti}}:\ti \in \setindex(\ci))$ , Eq.~\eqref{mean_r}
	\label{op:class_center}
	\State \textbf{end}
	\State \textbf{return} $\P_{\Rm}^{(\ci)}$ 
	\end{algorithmic}
\end{algorithm}
\begin{algorithm}
\caption{Minimum Distance to Riemannian Mean}
\label{alg:mdrm}
	Inputs: $\P_{\Rm}^{(\ci)}$, $\dK$ centers of classes from Algorithm~\ref{alg:r_mean}. \\
	Input: $\Y \in \Re^{\dF \dc\times\dt}$, an unlabelled test trial. \\
	Output: $\clout$, the predicted label of $\Y$.
	\begin{algorithmic}[1]
	\State Compute covariance matrix $\cov{}$ of $\Y$ 
	\label{op:cov}
	\State $\clout = \argmin_{\ci} \distR(\cov{}, \P_{\Rm}^{(\ci)})$
	\label{op:decision}
	\State \textbf{return} $\clout$ 
	\end{algorithmic}
\end{algorithm}

\subsection{Curve-based Online Classification} 

\begin{algorithm}
\caption{Curve-based Online Classification}
\label{alg:online}
	Inputs: hyper-parameters $\ws$, $\deltaN$, $\dd$, and $\pthres$.\\
	Inputs: $\P_{\Rm}^{(\ci)}$, $\ci = 1, \ldots, \dK$, centers of classes from Algorithm~\ref{alg:r_mean} (offline training). \\
	Inputs: Online EEG recording $\onlineX(\si)$. \\
	Output: $\cloutOn(\si)$, online predicted class.
	\begin{algorithmic}[1]
	\State $\di=1$
	\State \textbf{for} $\si = \ws$ \textbf{to} $\dr$ \textbf{step} $\deltaN$
	\State \quad Epoch $\Xs{\di}$, Eq.~(\ref{eq:online_epoch}), and classify it with Algorithm~\ref{alg:mdrm}
	\label{op3:epoch_and_classify}
	\State \quad \textbf{if} $\di \geq \dd$ 
	\label{op3:test_len}
	\State \quad \quad Find the most recurrent class in $\clarray = \clout_{\tj \in \Dindex(d)}$:
	$\cloutmp = \argmax_{\ci} \pocc(\ci)$ , Eq.~\eqref{eq:occ_prob}
	\label{op3:get_rho}
	\State \quad \quad \textbf{if} $\pocc(\cloutmp)>\pthres$
	\label{op3:test_rho}
	\State \quad \quad \quad Compute $\disvec_{\cloutmp}$ , Eq.~\eqref{eq:dist_vec} 
	\label{op3:get_disvec}
	\State \quad \quad \quad \textbf{if} $\disvec_{\cloutmp} < 0$ 
	\label{op3:test_disvec}
	\State \quad \quad \quad \quad \textbf{return} $\cloutOn = \cloutmp$
	\label{op3:valid_class}
	\State \quad \quad \quad \textbf{end}
	\State \quad \quad \textbf{end}
	\State \quad \textbf{end}
	\State \quad $\di = \di+1$
	\label{op3:increment_d}
	\State \textbf{end}
	\end{algorithmic}
\end{algorithm}


In offline synchronous BCI paradigm, cue onset are used as reference for the localization of a brain response, e.g. for an evoked potential. 
Nonetheless most of the BCI applications are online and asynchronous; cue onsets are not known, thus designing online version of BCI algorithms, even well documented ones, is not a trivial task. 
The approach introduced here identifies a period (\textit{i.e.} interval) in the online EEG $\onlineX \in \Re^{\dF\dc\times\dr}$, where $\dr$ is the number of recorded samples, associated with a high probability (above threshold) of observing an SSVEP at a specific frequency, as illustrated in Algorithm~\ref{alg:online}. 
  
To locate this interval, we focus on the last $\dd$ recorded EEG overlapping epochs 
$ \left\{ \Xs{\ensuremath{i}} \in \Re^{\dF\dc \times \ws} \right\}_{\tj \in \Dindex(d)} $ , 
with the set of indices $\Dindex(d) = \di-\dd +1,\ldots,\di-1,\di$; 
where $\di$ is the index of the current epoch $\Xs{\di}$ in the online recording $\onlineX(n)$. 
Epochs have size $\ws$, and the interval between two consecutive epochs is $\deltaN$, with $\ws > \deltaN $:
\begin{equation}
\label{eq:online_epoch}
 \Xs{\di} = \onlineX(\si-\ws,\ldots,\si) \ .
\end{equation}
To obtain the first $\dd$ epochs $\Xs{\ensuremath{\tj \in \Dindex(d)}}$, at least $\ws+(\dd-1)\,\deltaN$ samples of $\onlineX$ should be recorded (step~\ref{op3:test_len}).

The classification outputs $\clout_{\tj \in \Dindex(d)}$ obtained by applying Algorithm~\ref{alg:mdrm} (step~\ref{op3:epoch_and_classify})
on $\Xs{\ensuremath{\tj \in \Dindex(d)}}$ are stored in a vector $\clarray$. 
The class that occurs the most in $\clarray$ (step~\ref{op3:get_rho}), with an occurrence probability $\pocc(\ci)$ above a defined threshold $\pthres$
, is considered to be the class, denoted $\cloutmp$, of the ongoing EEG recording $\onlineX(\si)$. Vector $\pocc$ is defined as:
\begin{equation}
\label{eq:occ_prob}
\begin{split}
  \pocc(\ci) = \frac{\#\{ \clout_{\tj \in \Dindex(d)} = \ci \}}{\dd} \ & , \text{ for } \ci = 1, \ldots, \dK, 
	\\
  \text{and} \ \ \ \ \ \ \ \ \ \ \ \ \ \ \ \ \cloutmp = {\argmax}_{\ci} & \ \pocc(\ci) \ .
\end{split}
\end{equation}

If $\pthres$ is not reached within the last $\dd$ epochs, the classification output is held back, and the sliding process continues until $\pthres$ is reached. 
In the last $\dd$ epochs, once a class $\cloutmp$ has been identified,  a curve direction criterion is introduced to enforce the robustness of the result.  
For class $\cloutmp$ to be validated, this criterion requires that the direction taken by the displacement of covariance matrices $\cov{\textit{\tj} \in \Dindex(\textit{\di})}$ be toward the center of class $\P_{\Rm}^{(\cloutmp)}$. 
Hence $\disvec_{\cloutmp}$, the sum of gradients (\textit{i.e.} differentials) of the curve made by distances from $\cov{ \textit{\tj} \in \Dindex(\textit{\di})}$ to $\P_{\Rm}^{(\cloutmp)}$ should be negative (step~\ref{op3:test_disvec}):
\begin{equation}
\label{eq:dist_vec}
\begin{split}
  \disvec_{\cloutmp} &= \sum_{\tj \in \Dindex(\di)} \frac{\Delta \diffdist_{\cloutmp}(\tj)}{\Delta \tj} 
  \ = \sum_{\tj=\di-\dd+2}^{\di} \diffdist_{\cloutmp}(\tj) - \diffdist_{\cloutmp}(\tj-1)
	\ < 0 
	\\
  & \text{with} \ \ \ \ \diffdist_{\cloutmp}(\tj) = \frac{\distR (\cov{\textit{\tj}}, \P_{\Rm}^{(\cloutmp)} )}{\sum_{\ci=1}^{\dK} \distR (\cov{\textit{\tj}},\P_{\Rm}^{(\ci)})} \ .
\end{split}
\end{equation}


The occurrence criterion is inspired by the dynamic stopping of \cite{verschore2012dynamic}; there is no fixed trial length for classification.
The occurrence criterion ensures that the detected user intention is unaffected by any short time disturbances due to noise or subject's inattention, as presented in Algorithm~\ref{alg:online}. 
This approach offers a good compromise to obtain robust results within a short and flexible time.


\section{Experimental Validation}
\label{sec:exp_valid}
Covariance matrix estimators, Algorithms~\ref{alg:mdrm} and \ref{alg:online} are applied to SSVEP signals for offline and online analysis. This section presents the analysis and results obtained. 

\subsection{Data Description}
The signals are recorded from 12 subjects during an SSVEP experiment. 
EEG are measured on $\dc=$ 8 channels: $O_Z$, $O_1$, $O_2$, $PO_Z$, $PO_3$, $PO_4$, $PO_7$, and $PO_8$. 
The ground and the reference electrodes were placed respectively on $F_Z$ and the right hear mastoid. 
The acquisition rate is $T_s = 256$ Hz on a gTec MobiLab amp.
The subjects are presented with $\dF=$ 3 visual target stimuli blinking respectively at $\text{freq}=$ 13Hz, 17Hz and 21Hz. 
It is a $\dK=$ 4 classes BCI setup made of the $\dF=$ 3 stimulus classes and one resting class (no-SSVEP).
In a session, which lasts 5 minutes, 32 trials are recorded: 8 for each visual stimulus and 8 for the resting class. 
The number of sessions recorded per subject varies from 2 to 5. 
Thus the longest EEG recorded for a single subject is 25 minutes or 160 trials.
The trial length is 6 seconds, that is $\dt=6 \times T_s = 1536$ samples. 
Since data are rearranged as detailed in \eqref{eq:ext_data}, trials $\X \in \Re^{\dF\dc \times \dt}$, where $\dF\dc = 24$ corresponding to 8 channels times 3 stimulus frequencies.
For each subject, a test set is made of 32 trials while the remaining trials (which might vary from 32 to 128) make up for the training set.

\subsection{Covariance Estimators Comparison}

\begin{figure*}[t!]
		\centering
		\begin{minipage}[b]{0.8\linewidth}
        \pgfimage[width=1\textwidth]{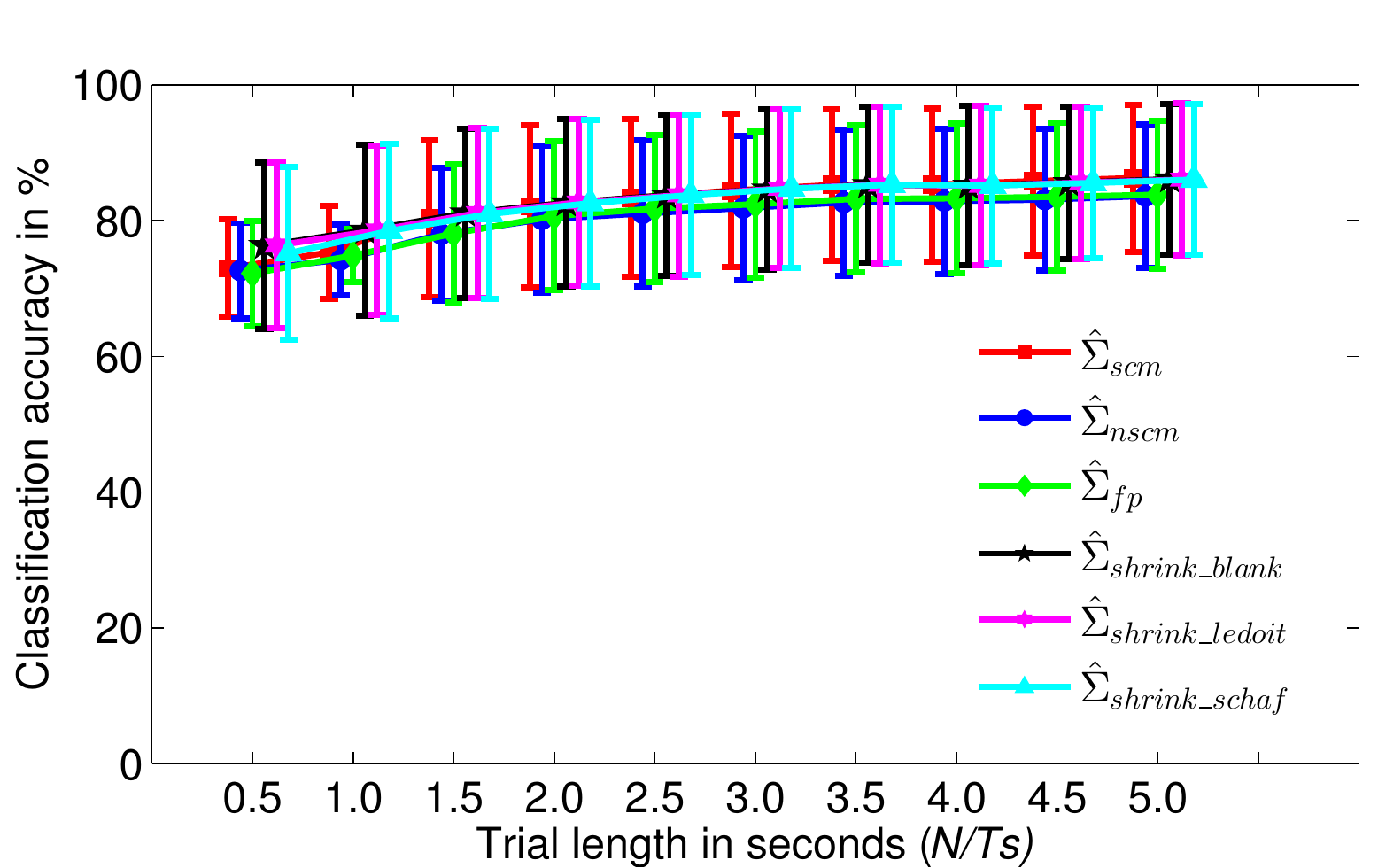}
        \subcaption{}
        \label{fig:acc_errorbar}
        \end{minipage}
        
        \begin{minipage}[b]{0.8\linewidth}
        \pgfimage[width=1\textwidth]{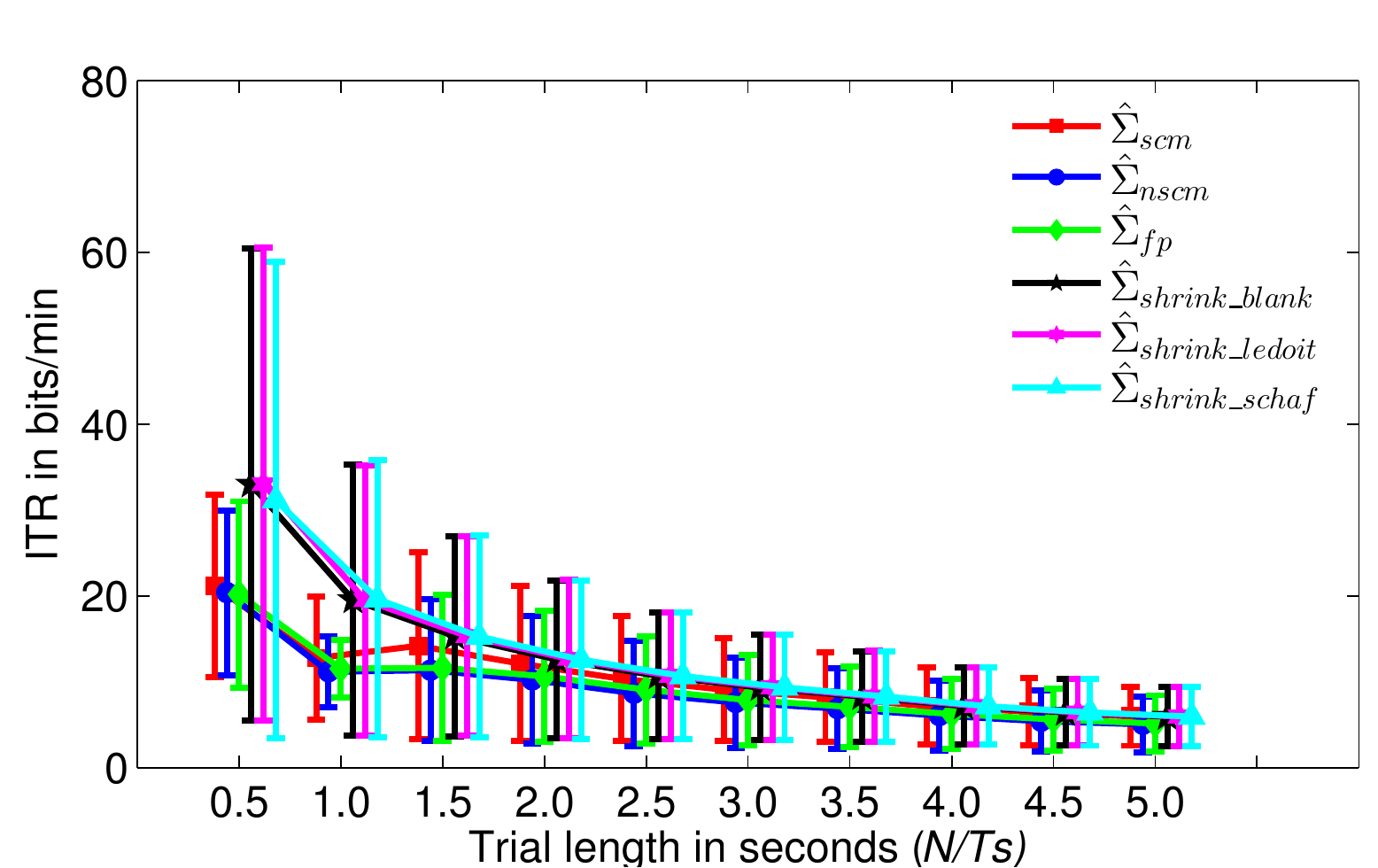}
        \subcaption{}
        \label{fig:itr_errorbar}
        \end{minipage}
        \caption{Comparison of covariance estimators in terms of classification accuracy  and Information Transfer Rate (ITR) obtained with MDRM with increasing EEG trial length. For each trial length, the average accuracy and ITR across all subjects and across all replications is shown. Bars indicate the standard deviation. (a) Accuracy in percentage, (b) ITR in bits/min.} 
\label{fig:class_size}
\end{figure*}

In this section, the effectiveness of covariance matrix estimators is evaluated for SSVEP signals. 
The evaluation is done in terms of classification accuracy, information transfer rate (ITR), and integrated discrimination improvement (IDI), obtained by each estimator (see Section~\ref{sec:cov_mat}) while using the offline MDRM classifier. 
The different conditioning of covariance matrices are also investigated.

A bootstrapping with 1000 replications is performed to assess the performances of each estimator. 
Estimators are compared on 10 trial lengths $t \in \{ 0.5, 1.0, \dots 5.0\}$ seconds, as these are known to affect the estimators performance. Here $\dt \in \{ 128, 256, \dots, 1280 \}$ is computed as $\dt=t \times T_s$.

Figure~\ref{fig:acc_errorbar} shows the classification accuracies of each estimator. 
The increase in the accuracy can be attributed to the fact that the relevant patterns in EEG accumulate with the trial length, producing better estimation of the covariance matrices. 
This is known to be particularly true for the SCM estimator and it could be seen in Figure~\ref{fig:acc_errorbar}.
It appears that shrinkage estimators (especially Ledoit and Sch\"afer) are less affected by the reduction of epoch sizes than the other estimators. This is a direct consequence of the regularization between the sample covariance matrices and the targeted (expected) covariance matrix of independent variables.  
The information of Figure~\ref{fig:acc_errorbar} is congruent with Figure~\ref{fig:itr_errorbar}; ITR increases as the trial length decreases. The transfer rate is higher for shrinkage estimators with shorter epoch sizes, while maintaining correct accuracy.

\begin{figure*}[ht!]
		\centering
		\begin{minipage}[b]{0.8\linewidth}
        \pgfimage[width=1\textwidth]{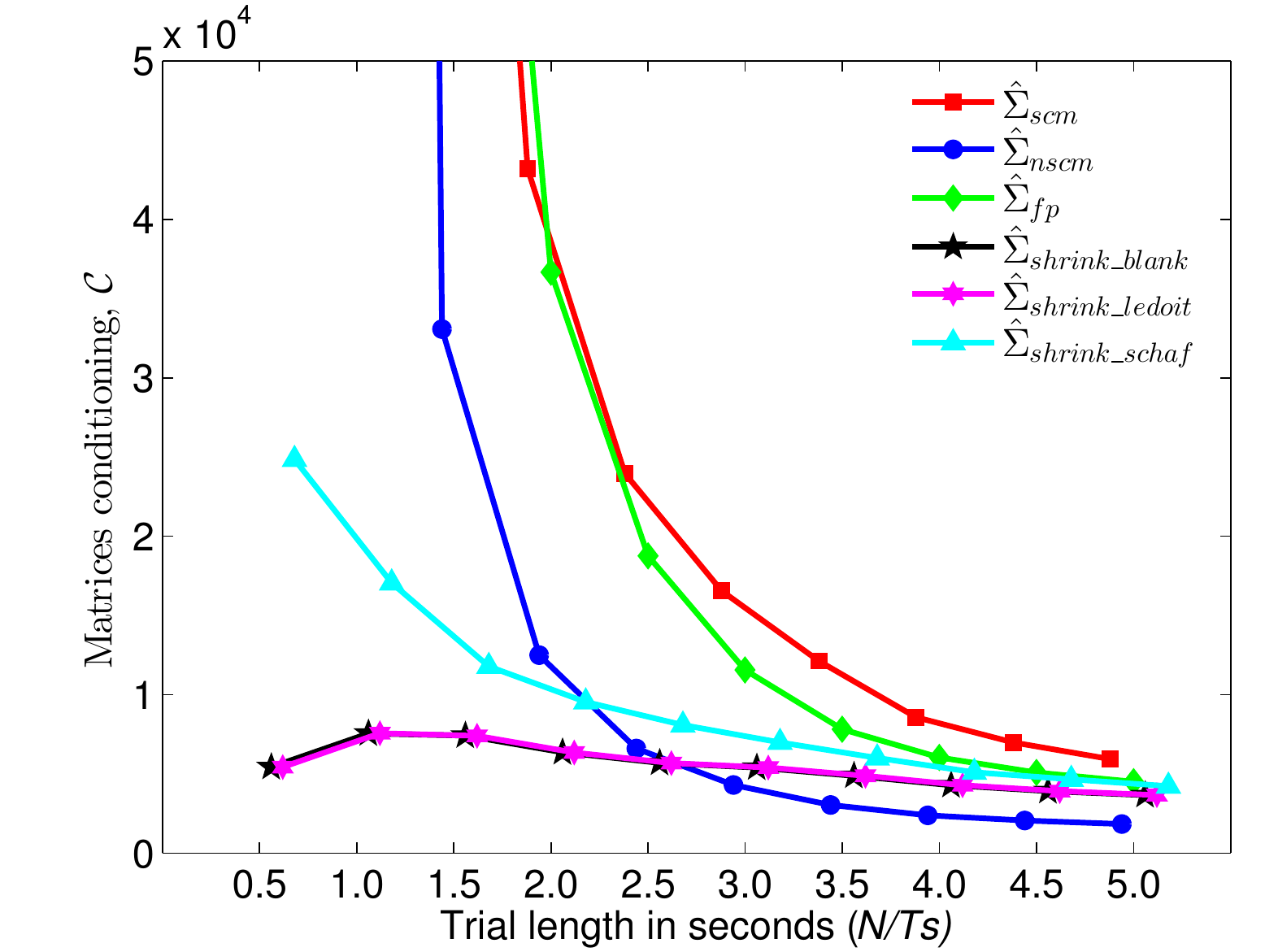}
        \subcaption{}
        \label{fig:eigenvalue_range}
        \end{minipage}
        
        \begin{minipage}[b]{0.8\linewidth}
        \pgfimage[width=1\textwidth]{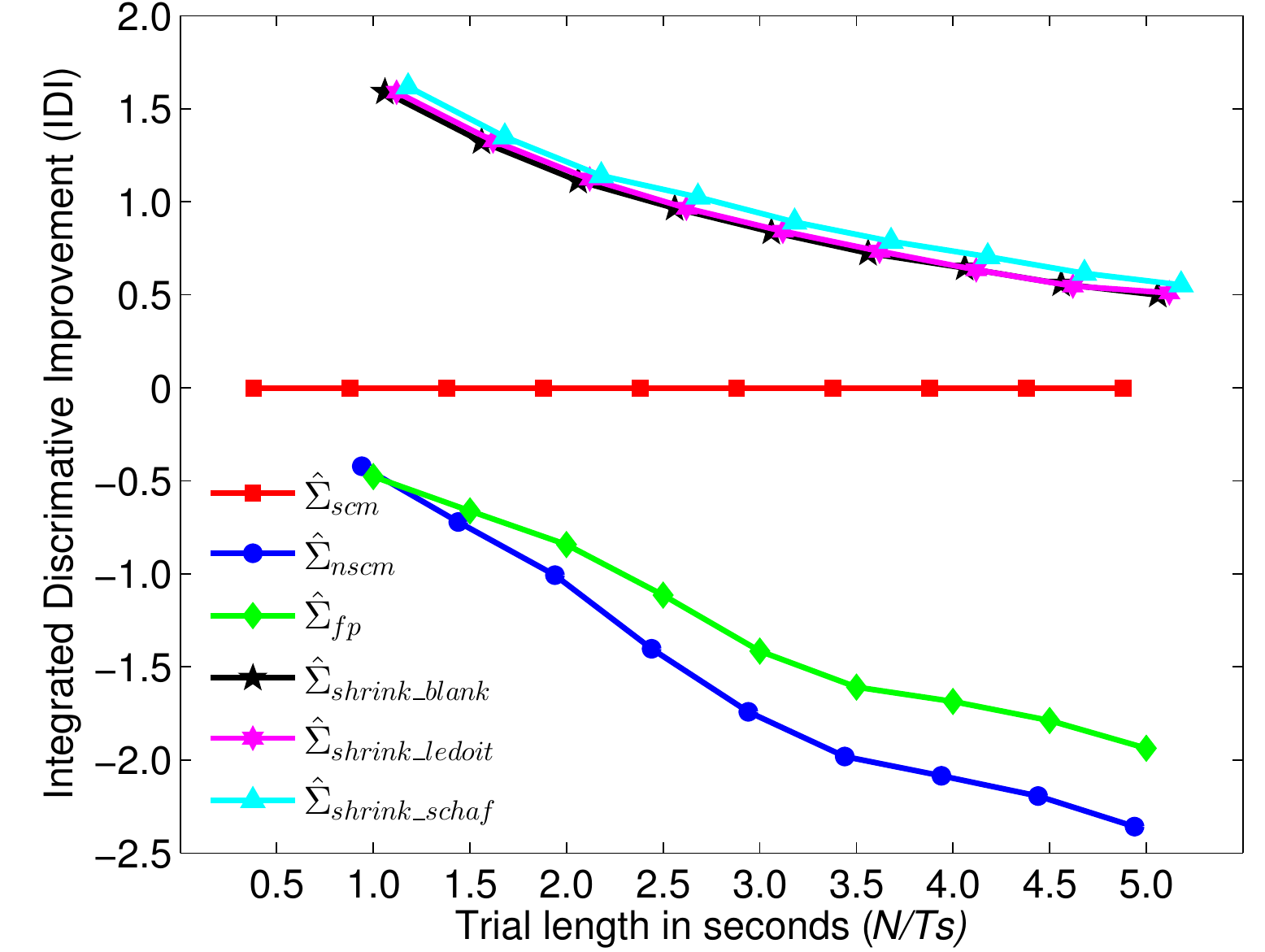}
        \subcaption{}
        \label{fig:idi}
        \end{minipage}
        
        \caption{(a) Covariance matrices condition expressed as the ratio $\mathcal{C}$ between largest and smallest eigenvalues for the different covariance estimators. The comparison is done for increasing EEG trial length. (b) Integrated discrimination improvement brought to the classification task by various estimators along varying trail length. The indicated IDI values are multiplied by $10^{2}$. $\cov{scm}$ is used as baseline. }
        \label{fig:class_idi}
\end{figure*}

For computational purposes, it is important to look at the matrix conditioning. 
Figure~\ref{fig:eigenvalue_range} shows the ratio $\mathcal{C}$ between the largest and smallest eigenvalues: in well-conditioned matrices, $\mathcal{C}$ is small. 
Shrinkage estimators offer better conditioned matrices while the SCM, NSCM, and Fixed Point matrices are ill-conditioned below 2 seconds of trial length, and may result in singular matrices. 

On Figure~\ref{fig:idi}, the Integrated Discrimination Improvement (IDI), as defined in \cite{pencina2008evaluating}, is computed for the different estimators and trial lengths. 
The SCM is used as a reference for improvement, as this is the most popular estimator in the literature. 
Negative IDI means a deterioration in the method discrimination ability. 
It is clear that shrinkage estimators increase the discrimination power of the classifier. 
However, despite being more complex than the SCM, the NSCM and the Fixed Point estimators decrease the discrimination ability of classifiers.     
From Figures~\ref{fig:acc_errorbar} and~\ref{fig:idi}, it is apparent that the difference in performance between the SCM and shrinkage estimators reduces as the trial length increases.
The simplicity of the SCM plays a favorable role: it is an attractive method for longer trials. 
However for trial lengths compatible with a reasonable ITR for man-machine interaction ($< 5$ sec), the shrinkage estimators offer better performances, with the Schafer estimator being the best. 
The $p$-values under the hypothesis that there is no improvement (\textit{i.e.} IDI = 0) from one estimator to another are all inferior to $10^{-47}$, hence the improvement is significant.
It should be noted that the estimation of covariance matrices is a trade-off between the quality of the estimate and the computation time required; this should be considered for real-time processing.
    
\subsection{Effect of Outliers on Centers Estimation}

\begin{figure*}[ht!]
		\centering
		\begin{minipage}[b]{0.49\linewidth}
        \pgfimage[width=1\textwidth]{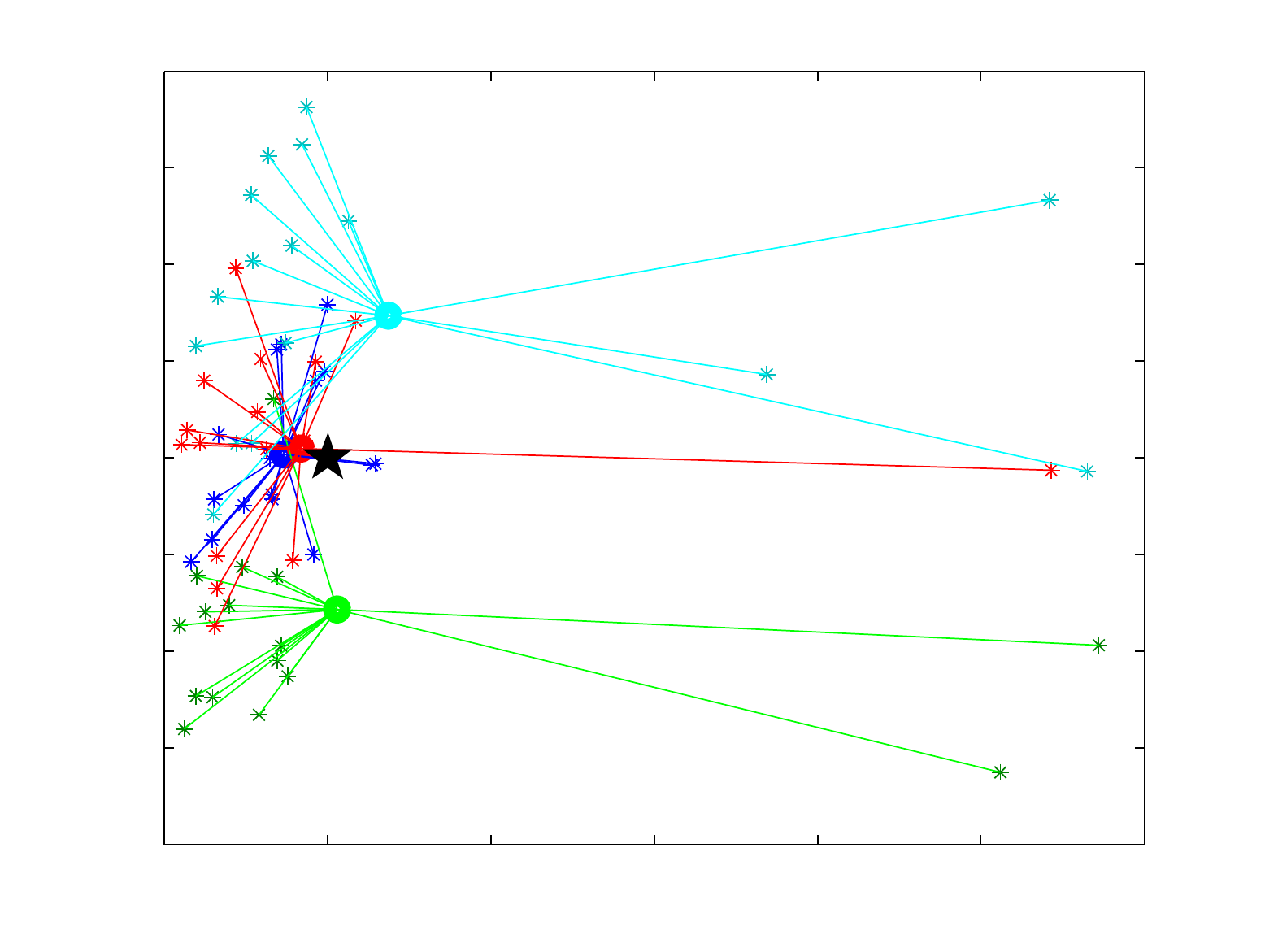}
        \subcaption{}
        \label{fig:tan_plan_s16_a}
        \end{minipage}
        \begin{minipage}[b]{0.49\linewidth}
        \pgfimage[width=1\textwidth]{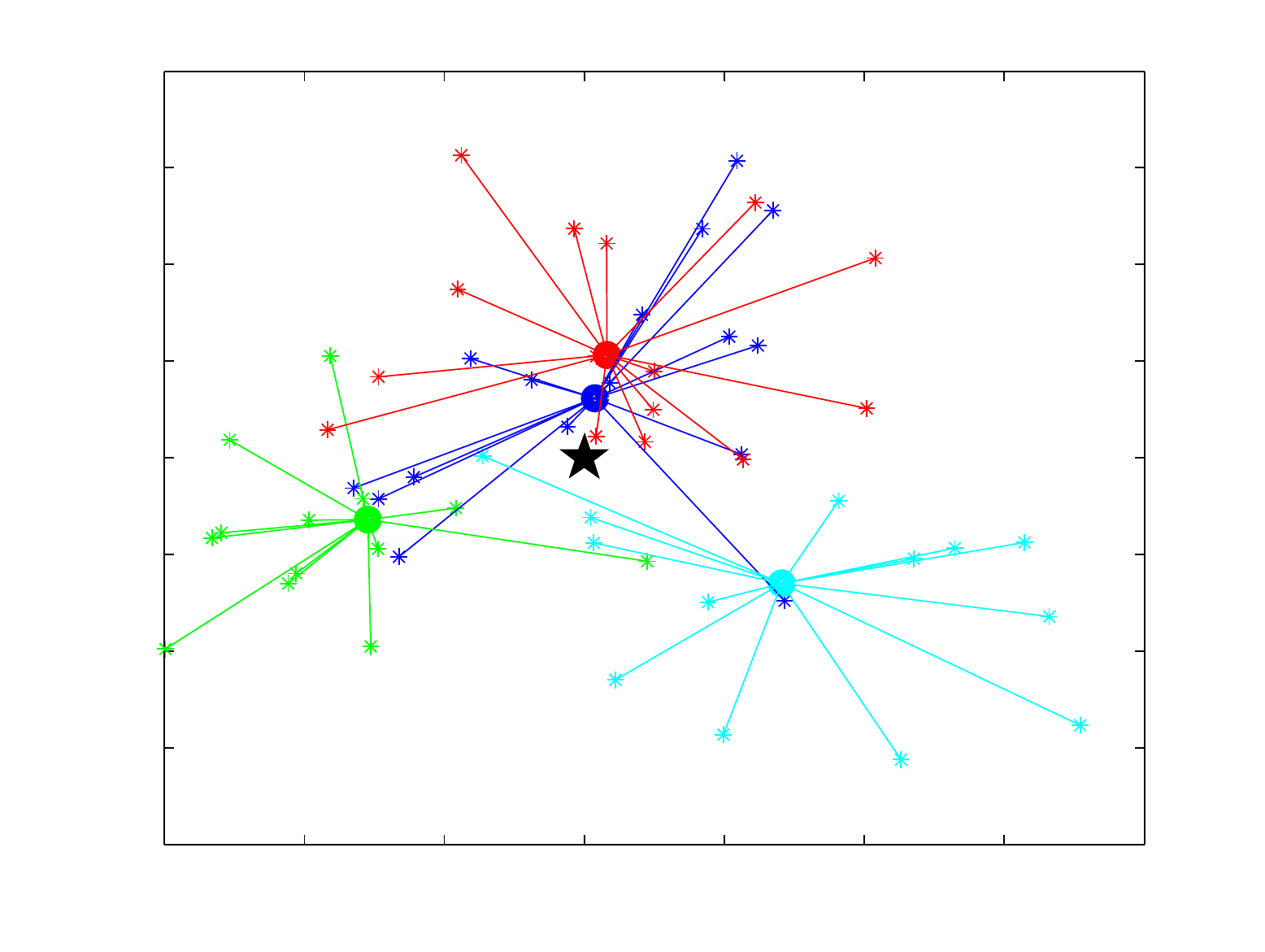}
        \subcaption{}
        \label{fig:tan_plan_s16_b}
        \end{minipage}
        \begin{minipage}[b]{0.49\linewidth}
        \pgfimage[width=1\textwidth]{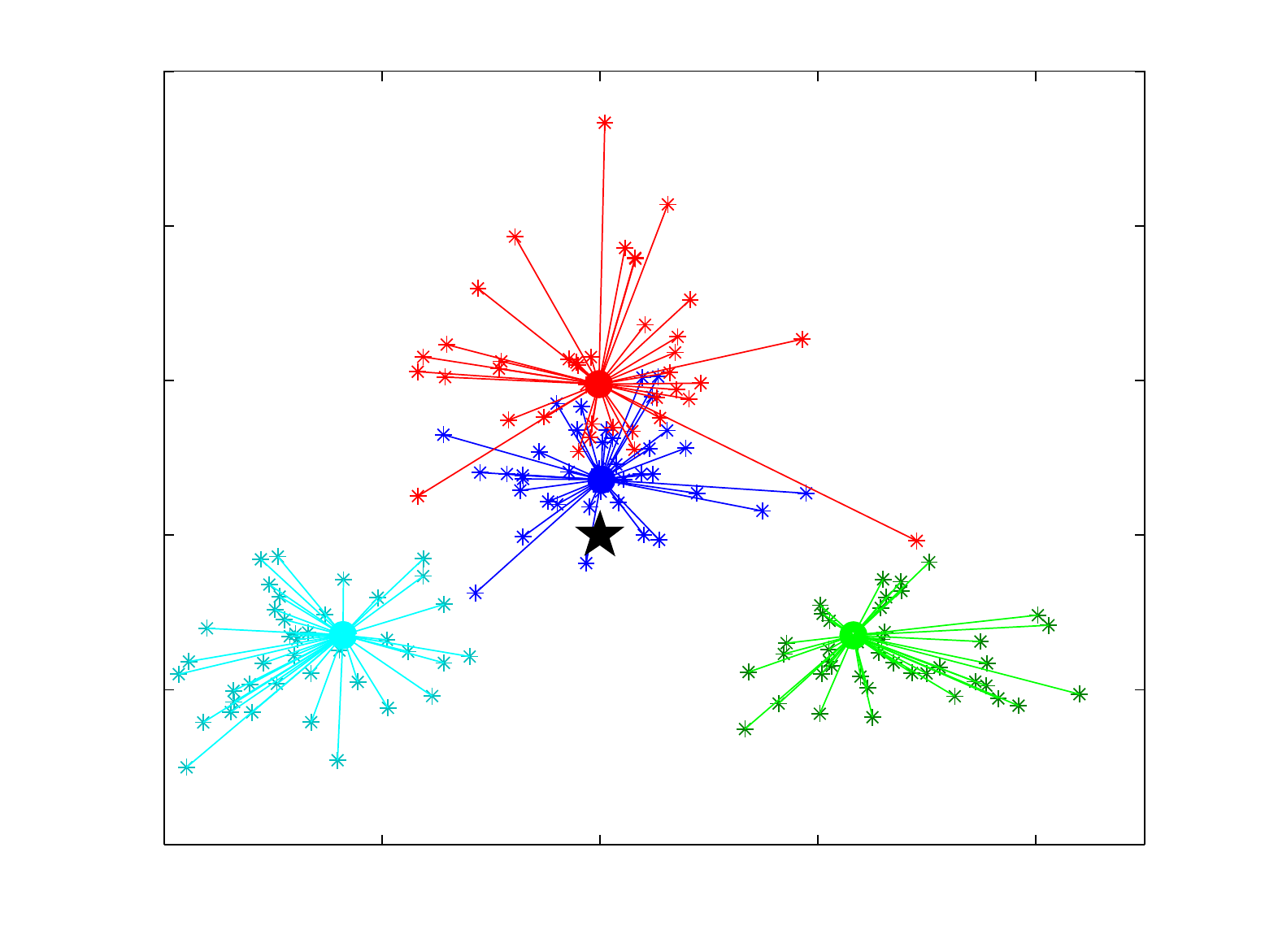}
        \subcaption{}
        \label{fig:tan_plan_s17_a}
        \end{minipage}
        \begin{minipage}[b]{0.49\linewidth}
        \pgfimage[width=1\textwidth]{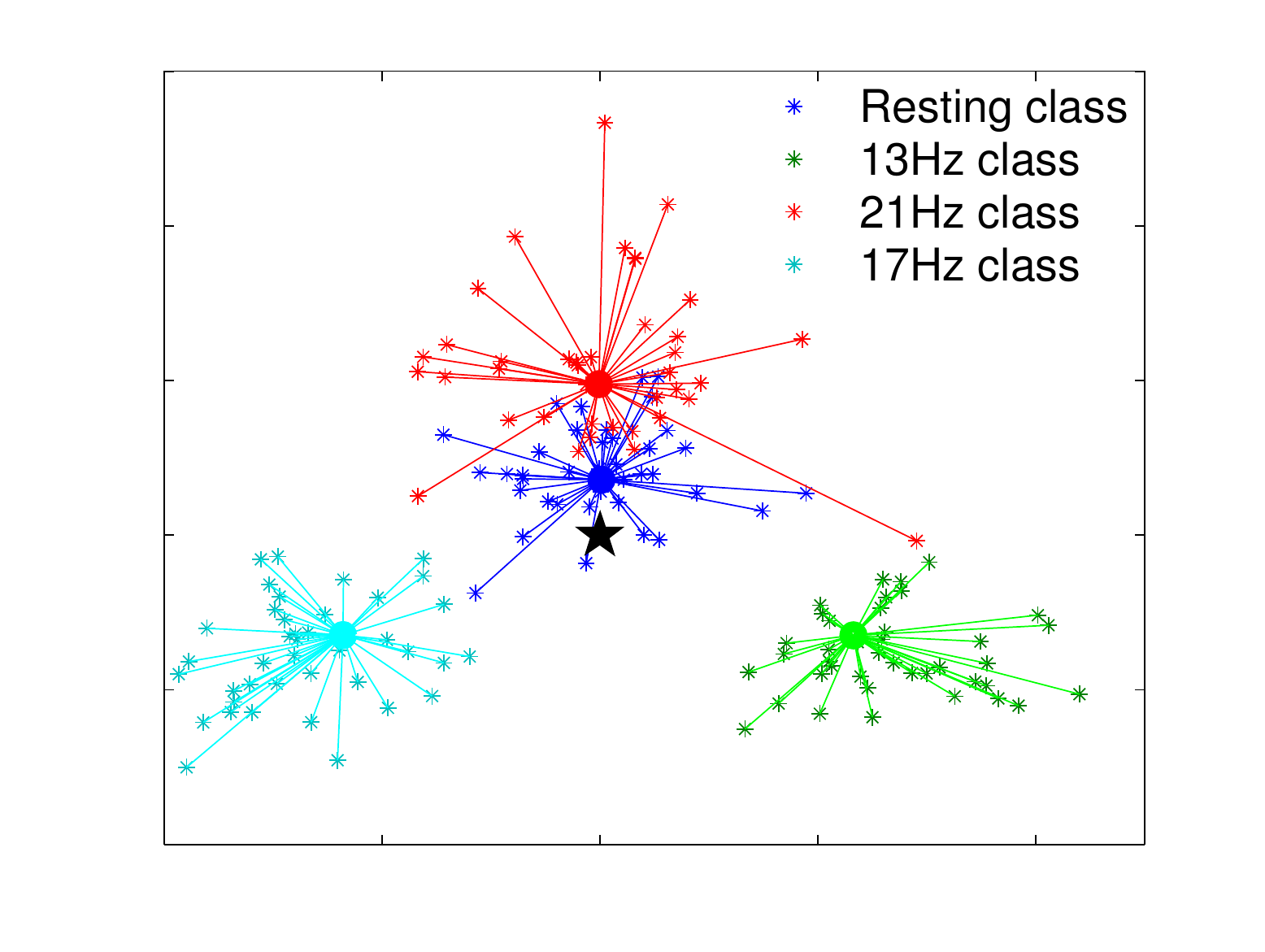}
        \subcaption{}
        \label{fig:tan_plan_s17_b}
        \end{minipage}
        \caption{Scatter plot of covariance matrices for all trials mapped on the tangent space. The distance between each trial covariance matrix $\P_{\ti}$ and its Riemannian mean class $\P_{\Rm}^{(\ci)}$ is shown as connection line. The black star represents the Riemannian mean of all trials. 
Subject with lowest BCI performance, (a) before and (b) after Riemannian potato filtering. 
Subject with highest BCI performance, (c) before and (d) after Riemannian potato filtering.}
\label{fig:tan_plan_s16&17}
\end{figure*}

Outliers can affect the offline training of the $\dK$ centers of class $\P_{\Rm}^{(\ci)}$ by Algorithm~\ref{alg:r_mean}.
Figure~\ref{fig:tan_plan_s16&17} shows representations of training covariance matrices $\P_{\ti}$ in the tangent space ($\S_{\ti}$), projected at the mean of all training trials (black star).
To obtain this visualization, the first two principal components of a PCA applied on $\left\{ \S_{\ti} \right\}_{\ti=1}^{\dT}$ are selected.
In Figures~\ref{fig:tan_plan_s16_b} and~\ref{fig:tan_plan_s17_b}, a Riemannian potato~\cite{barachant2013riemannian} is applied with the Riemannian mean of all training matrices as a fixed reference. 
Riemannian potato only removes highly corrupted signals, allowing to remove outlier trials that negatively impact the estimation of class centers.
Applying this filter in an offline evaluation improves estimation of class centers for the subject with the lowest BCI performance.
Figure~\ref{fig:tan_plan_s16_b} illustrates how center estimates are affected by the filtering. Bad estimates of a class center will result in poor clustering.  

It should be noted that the Riemannian potato may have a negative effect if the different classes are well separated.
The Riemannian potato could reject trials that are close to the average of their class but that are far from the mean of all trials.
Thus, ``legitimate'' trials could be discarded, even if they have a positive impact on the classification results.
Avoiding this central bias is a possible improvement of the Riemannian potato algorithm. 

\subsection{Classification Results}

In this section the classification performances are presented. It is divided into two parts: first, the relevance of identifying the latency between cue onset and SSVEP response is shown through the evaluation of the impact of SSVEP delays considered for offline classification. Then, performances of offline and online algorithms are compared and analysed. 

\subsubsection{SSVEP Response Time Delay for Offline}
\label{sec:ssvep_response_delay}

\begin{figure*}[ht!]
		\centering
		\begin{minipage}[b]{0.9\linewidth}
        \pgfimage[width=1\textwidth]{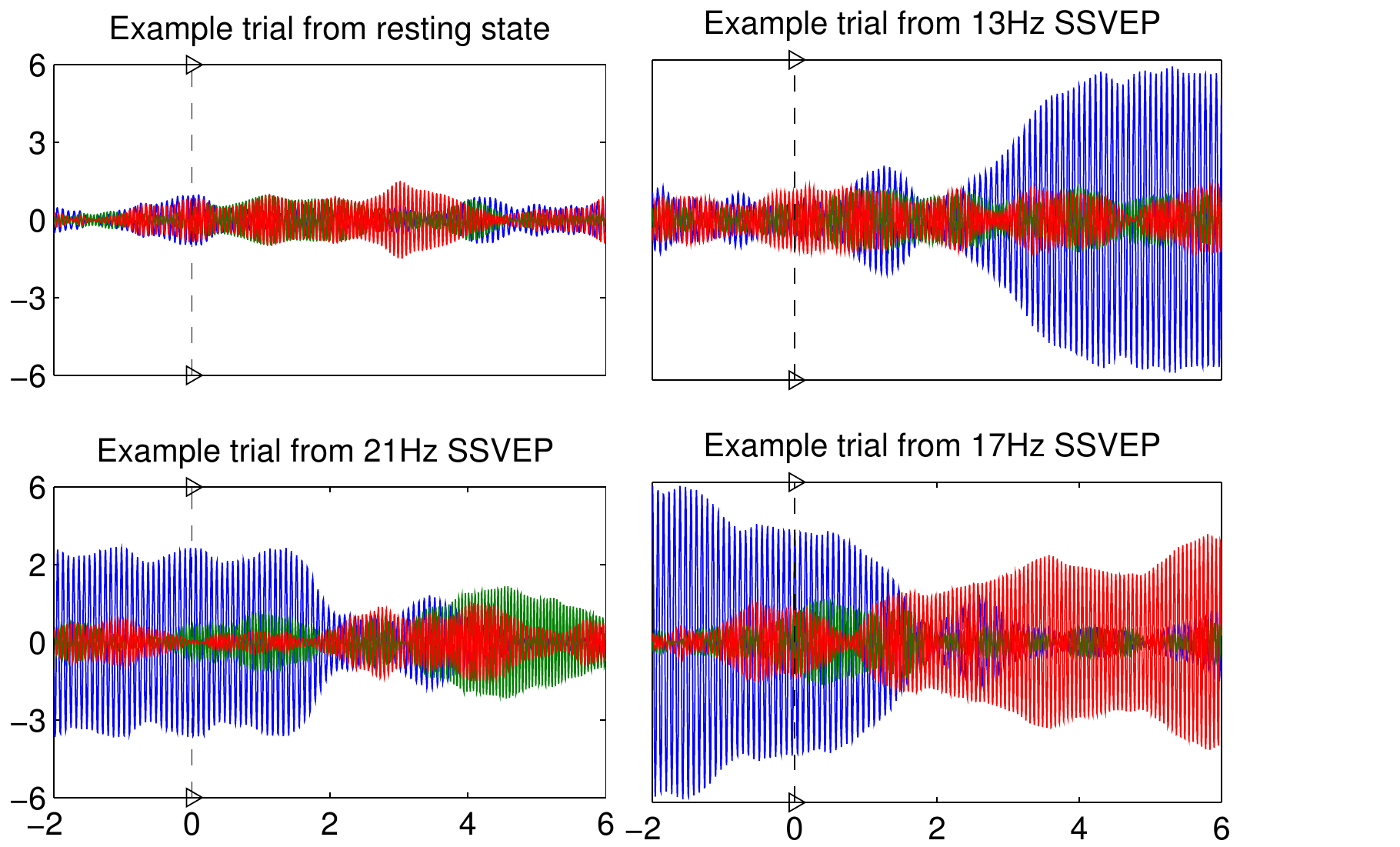}
        \subcaption{Example of trials for the subject with the highest BCI performance}
        \end{minipage}
        
		\begin{minipage}[b]{0.9\linewidth}
        \pgfimage[width=1\textwidth]{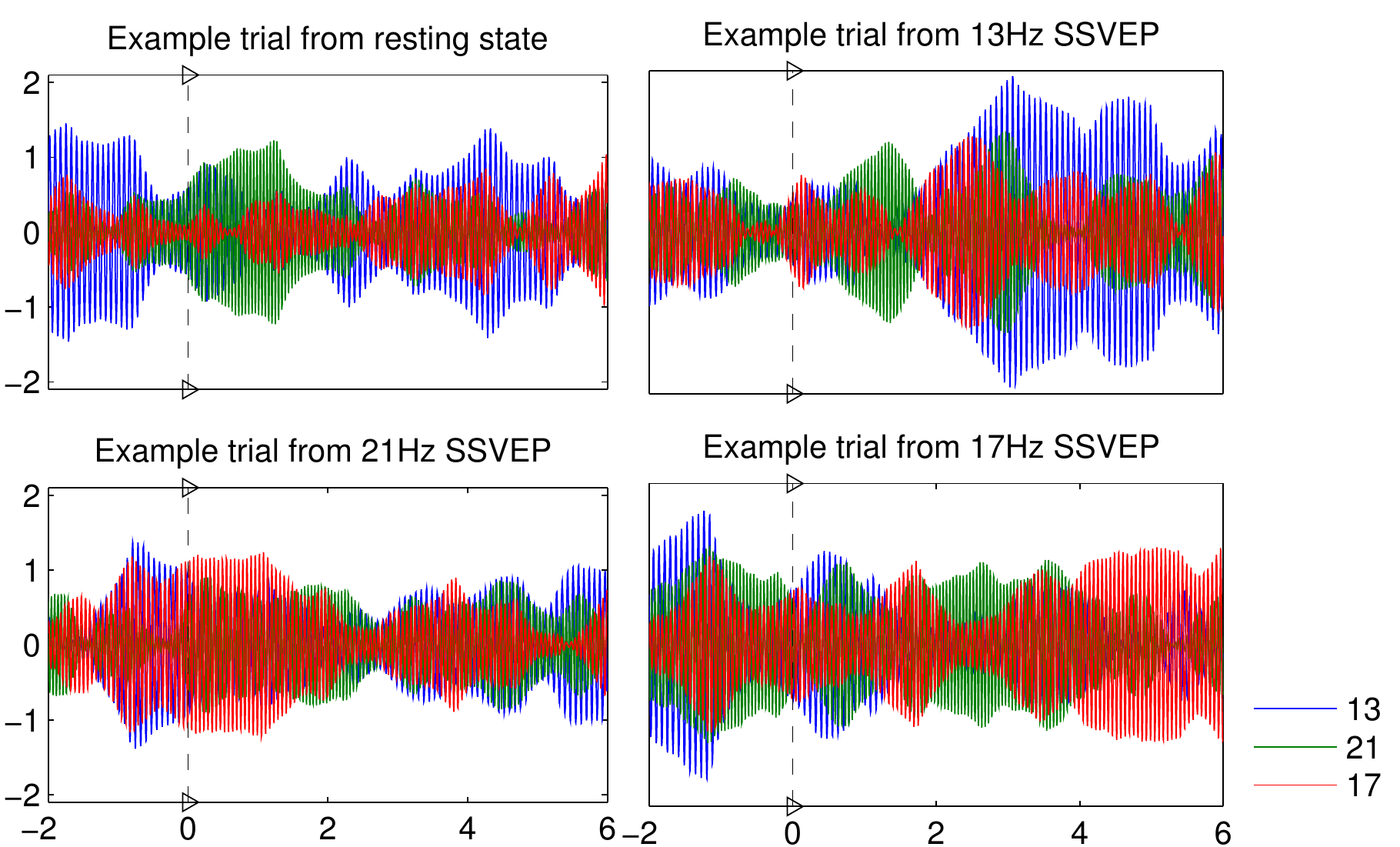}
        \subcaption{Example of trials for the subject with the lowest BCI performance}
        \end{minipage}
        \caption{Signal amplitude at each stimulus frequency, showing synchronization of EEG with respect to time (seconds). The raw signal of the trial measured on Oz is band filtered using a Butterworth of order 8 at each stimulus frequency and the resulting signals are shown in blue (dark grey) , green (grey), and red (light grey) for the same signal filtered respectively at 13, 17, and 21 Hz. The cue onset $\cue$ at time $0$ on the x-axis is shown with a vertical discontinued line. 4 trials are shown, one for each class. Signals from the subjects, (a) with highest BCI performance and (b) with lowest BCI performance.}
        \label{fig:sync_delay}
\end{figure*}

Figure~\ref{fig:sync_delay} shows time signals of one trial from each class, measured at $Oz$, for the subject with highest BCI performance and for the subject with lowest BCI performance. 
The figure displays the raw signal filtered at the 3 different stimulus frequency, each with a different color: 13Hz in blue, 17Hz in red, 21Hz in green.
It is visible that synchronization for current trial frequency appears only 2 seconds after cue onset $\cue = 0 s$ on the $x$-axis of the plots.
In fact, before $\cue + 2$ seconds the signal is still synchronized with the previous trial frequencies. 
In the signals from the subject with the highest BCI performance, synchronization at the trial stimulus frequency is predominant and permanent after $\cue + 2$, which eases the classification task. 
Moreover, for the resting trial (no SSVEP), the power in all stimuli frequencies is sensibly decreased. 
However, the poor predictive capability for the subject with lowest BCI performance could be explained by the fact that the predominance of the trial stimulus frequency is not always evident and not permanent. 
Also, during the resting trial, the power in stimuli frequencies is not significantly decreased.    

\begin{figure*}[ht!]
		\centering
		\begin{minipage}[b]{0.8\linewidth}
        \pgfimage[width=1\textwidth]{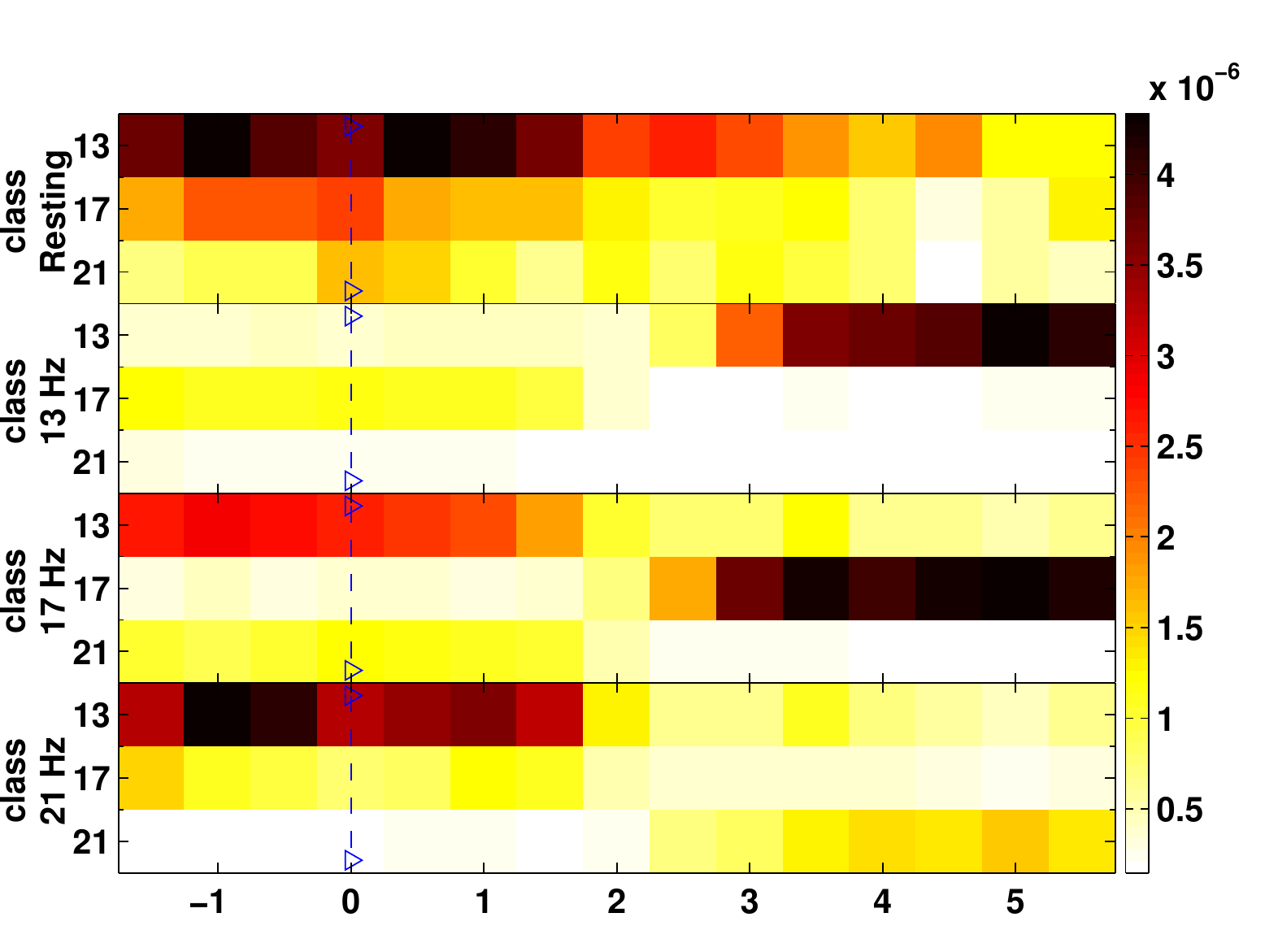}
        \subcaption{Time-frequency diagrams for the subject with the highest BCI performance}
        \end{minipage}
        
        \begin{minipage}[b]{0.8\linewidth}
        \pgfimage[width=1\textwidth]{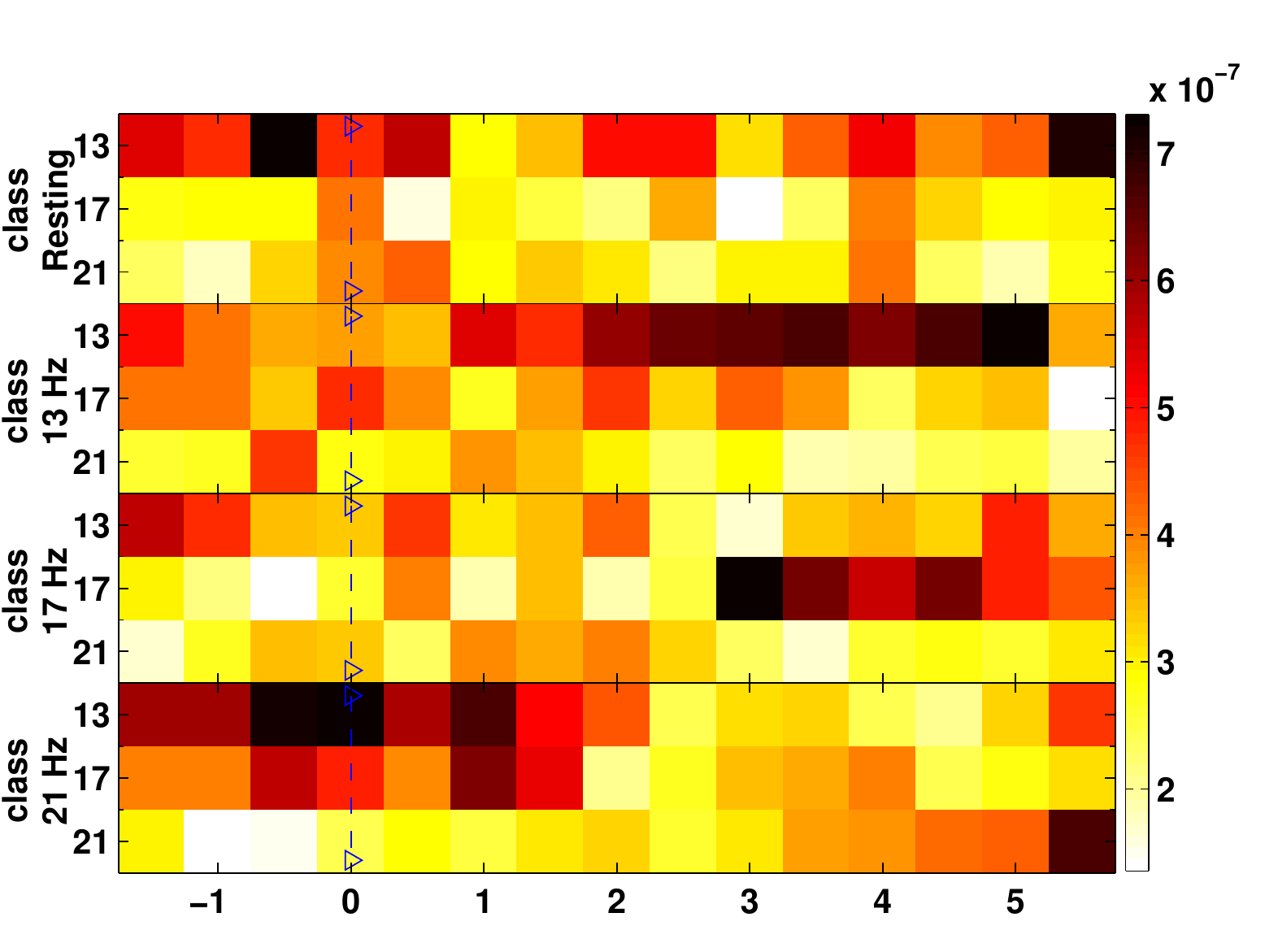}
        \subcaption{Time-frequency diagrams for the subject with the lowest BCI performance}
        \end{minipage}
        \caption{Time-frequency representation of EEG from various classes showing synchronization at in frequency bands (indicated on y axis) on both side of the cue onset. The power of synchronization is shown in the color bar. Dark colors show strong synchronization. 4 trials (top to bottom) are represented, one for each class. Signals from the subjects, (a) with highest BCI performance and (b) with lowest BCI performance.}
        \label{fig:sync_delay_psd}
\end{figure*}

Similar observation is made from Figure~\ref{fig:sync_delay_psd} where the time-frequency representations of EEG trials per class are plotted for the subjects with the highest and lowest BCI performances. 
The time-frequency representation is obtained with a STFT on rectangular windowing.
Each plot has 4 rows, one for each class, displaying the power spectral density (averaged over all trials) in the 13Hz, 17Hz, and 21Hz frequency bands as a function of the trial time.
An important increase in average classification accuracy (almost 10\%) could be obtained by taking the trial from 2 seconds after cue onset.
It is therefore crucial to consider the latency between the trial's cue onset and the actual synchronization of SSVEP at stimulus frequency. 
 
\subsubsection{Online Classification Results}
In the offline synchronous analysis, the confident window for classification was found to be at least 2 seconds after the cue onset ($\cue+2$). 
In an online asynchronous experiment, there is no cue onset, and the delay before SSVEP synchronization might defer from a trial to another and from a subject to another. 
To locate the trust EEG region for the classification, $\dd$ and $\pthres$ are set respectively to 5 and 0.7 through cross validation. The epoch size is $\ws = 3.6 $ seconds, and the overlap is $\deltaN = 0.2$ second.

\begin{figure*}[ht!]
		\centering
		\begin{minipage}[b]{0.8\linewidth}
        \pgfimage[width=1\textwidth]{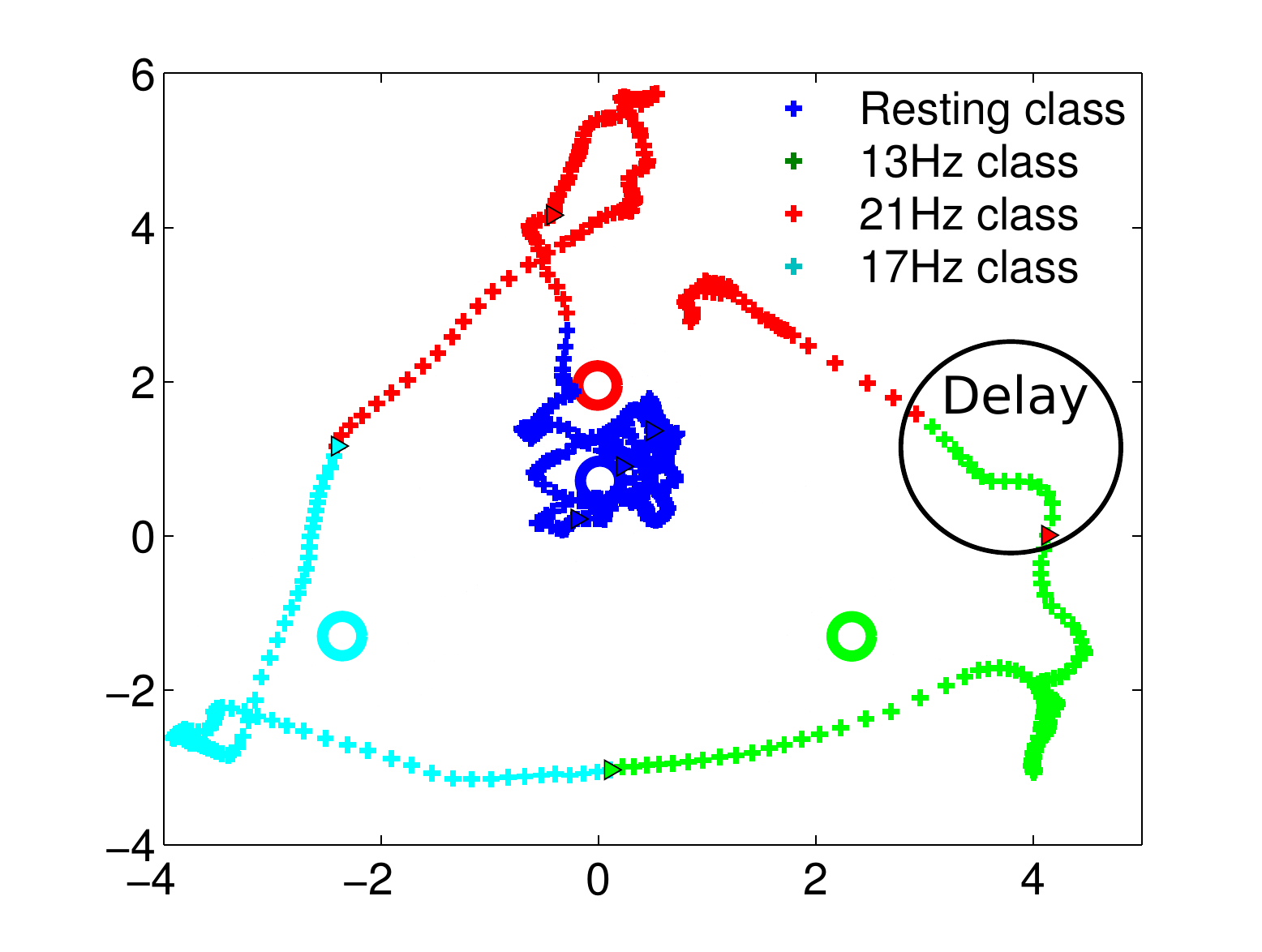}
        \subcaption{ }
        \label{fig:class_path1_s17}
        \end{minipage}      
        \begin{minipage}[b]{0.8\linewidth}
        \pgfimage[width=1\textwidth]{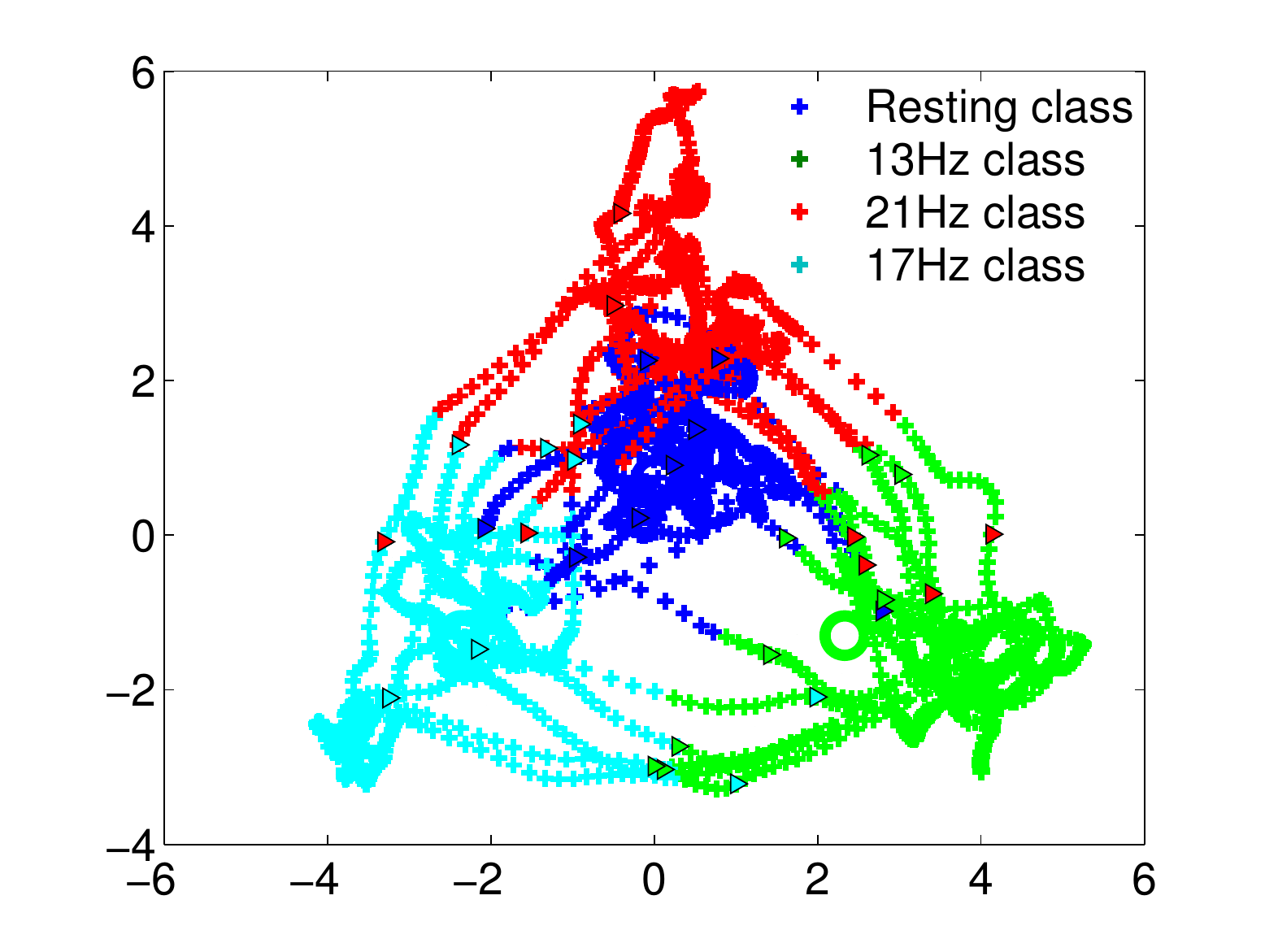}
        \subcaption{ }
        \label{fig:class_path2_s17}
        \end{minipage}
        
        \caption{Covariance matrices trajectory during a 4-class SSVEP online recording. The circles represent class centers. The triangles mark the beginning in the experiment of a new trial whose class is indicated by the triangle's color. \ref{fig:class_path1_s17} shows the first 7 trials. The first 3 trials are from the resting class, the remaining are respectively class 13Hz, 17Hz, and 21Hz. \ref{fig:class_path2_s17} shows the entire recording. Data are from the subject with the highest BCI performance.}
\label{fig:class_path_s17}
\end{figure*}

The observation of Figure~\ref{fig:class_path_s17} provides a visualization of the principle guiding the online implementation of Equation~\eqref{eq:dist_vec}.
This figure shows the trajectory on the tangent space taken by covariance matrices during a 4-class SSVEP experiment, and how they are classified epoch by epoch. 
It can be seen (encircled in Figure~\ref{fig:class_path1_s17}) that a change in the SSVEP stimulus might not be detected instantaneously by the classifier. 
The trials are erroneously attributed with confidence to the previous class. 
The proposed online algorithm, described in Algorithm~\ref{alg:online}, mitigates this issue and increases the classification accuracy as shown in Table~\ref{tab:res}.  
In the `Offline' column are the classification accuracies for each subject obtained by the application of the state-the-art MDRM as described in Algorithm~\ref{alg:mdrm}, which classifies full trials. 
Column `Offline opt.' shows the results of this approach improved by taking into consideration the SSVEP response time delay as found in section~\ref{sec:ssvep_response_delay}: trials are taken 2 second after cue onset.      
Columns `Online' and `Online + $\disvec_{\cloutmp}$' contain the performance of the online algorithm as described in Algorithm~\ref{alg:online}.
The `Online' column shows the results of the online algorithm without the curve direction criterion (\textit{i.e.}, without steps 6 to 11), and `Online + $\disvec_{\cloutmp}$' shows the improvement brought by this criterion. 
The performances are in terms of average classification accuracy (acc(\%)) and average time taken into the trial before classification (delay(s)). 
For the offline approaches, classification is only done at the end of the trial \textit{i.e.} 6 seconds after cue onset.  
\\
The curve direction criterion increases the rejection of epochs that could be wrongly classified, it thus significantly increases the classification accuracy of the online algorithm (74.73\% to 79.13\%), while increasing the delay (1.006 s to 1.239 s) before classification. 
When compared to the state-of-the-art MDRM, the curve-based classification yields better results both in terms of classification accuracy and delay before classification, \textit{i.e.} from 71.51\% to 79.13\% and from 6 s (or longer, depending on trial length $\dt$) to 1.239 s.
There is a small drop in classification accuracy from optimized offline (81.27\%) to online (79.13\%) algorithms.
However, classification outputs are reached faster with the online algorithm.
Moreover, the online algorithm can be applied in both synchronous and asynchronous paradigms, while the offline algorithms are limited to synchronous paradigms which provide strongly limited user interaction. 
 
\begin{table*}[ht!]
\centering
\ra{1}
\begin{tabular}{c|c|c||c|c|c|c|}
\cline{2-7}
&  \multicolumn{1}{ c|}{Offline} &\multicolumn{1}{ c|| }{Offline opt.} & \multicolumn{2}{ c| }{Online} &\multicolumn{2}{ c| }{Online + $\disvec_{\cloutmp}$} \\ \cline{1-7}
\multicolumn{1}{ |c| }{Sub.} & acc(\%) & acc(\%) & acc(\%) & delay(s) & acc(\%) & delay(s)\\ \hline
\multicolumn{1}{ |c| }{1} & 62.50 & 73.44 & 70.31 & \textbf{1.081} & \textbf{71.37} & 1.338 \\ \hline
\multicolumn{1}{ |c| }{2} & 67.19 & 79.69 & \textbf{78.12} & \textbf{0.950} & 75.00 & 1.303 \\ \hline
\multicolumn{1}{ |c| }{3} & 85.94 & 85.94 & 87.50 & \textbf{0.950} & \textbf{89.06} & 0.972 \\ \hline
\multicolumn{1}{ |c| }{4} & 78.12 & 87.50 & 73.44 & \textbf{1.016} & \textbf{75.00} & 1.056 \\ \hline
\multicolumn{1}{ |c| }{5} & 70.31 & 68.75 & 70.31 & \textbf{1.013} & \textbf{70.31} & 1.319 \\ \hline
\multicolumn{1}{ |c| }{6} & 79.69 & 85.94 & \textbf{87.50} & \textbf{0.997} & 87.35 & 1.338 \\ \hline
\multicolumn{1}{ |c| }{7} & 78.12 & 88.54 & 84.38 & \textbf{1.004} & \textbf{85.42} & 1.144 \\ \hline
\multicolumn{1}{ |c| }{8} & 89.06 & 92.19 & 85.94 & \textbf{0.947} & \textbf{89.06} & 1.072 \\ \hline
\multicolumn{1}{ |c| }{9} & 62.50 & 70.31 & 67.19 & \textbf{0.922} & \textbf{75.00} & 1.244 \\ \hline
\multicolumn{1}{ |c| }{10} & 60.94 & 80.47 & 63.28 & \textbf{1.044} & \textbf{69.53} & 1.264 \\ \hline
\multicolumn{1}{ |c| }{11} & 43.75 & 65.62 & 59.38 & \textbf{1.109} & \textbf{68.75} & 1.375 \\ \hline
\multicolumn{1}{ |c| }{12} & 80.00 & 96.88 & 69.37 & \textbf{1.036} & \textbf{93.75} & 1.443 \\ \hline \hline
\multicolumn{1}{ |c| }{mean} & 71.51 & 81.27 & 74.73 & \textbf{1.006} & \textbf{79.13} & 1.239 \\ \hline
\end{tabular}
\caption{Average classification accuracies (acc(\%)) achieved using the state-of-the-art MDRM (Offline), the optimized version of the MDRM (Offline opt.), the online algorithm without the curve direction criterion (Online), and the complete online algorithm (Online + $\disvec_{\cloutmp}$). For the online algorithms, the delay between trial start and classification is also provided (delay(s)).}
\label{tab:res}
\end{table*}


\section{Conclusion}

This work investigated the efficiency of Riemannian geometry when dealing with covariance matrices as classification features. 
A novel algorithm, relying on the curve direction in the space of covariance EEG signals, was introduced and applied on a SSVEP classification task for a 4-class brain computer interface. 
Different covariance matrix estimators were investigated and their robustness was assessed on multichannel SSVEP signals to ensure that the obtained matrices are accurate estimates of data covariance, are well conditioned, and verify the positive-definiteness property. 
The Sch\"afer shrinkage estimator was found to be the best as it yielded the highest classification accuracy with the MDRM algorithm.

The introduced online algorithm enhances the stability of the BCI system, balancing between the classification speed and the prediction accuracy. 
Classification confidence over a number of epochs mitigates the short term perturbations in the experimental conditions and the attentional variations of the subject,
wile considering the curve direction overcomes the misclassification of EEG trials that are still synchronized with past stimuli frequencies at classification time. 

The Riemannian geometric approach  
and the determination of the Riemannian mean of classes, 
offers a powerful tool to develop robust algorithms.
A possible extension to the proposed algorithm is to include a dynamic adaptation of the Riemannian mean for each class.
This dynamic adaptation was implemented, using the latest successfully classified covariance matrices to replace to oldest ones in the computation of the mean.
It is not reported in this paper as there was no significant improvement in the results, which could be attributed to the relatively short duration of the EEG recording sessions, which were no longer than 25 min. 
The Riemannian means of classes do not change much over the sessions, but for longer recording time this adaptation could bring significant improvements. 

\section*{Acknowledgment}
The authors would like to thank Louis~Mayaud from Mensia Technologies for his contribution in discussions that led to the completion of this work.  

\bibliography{BCIBib}

\begin{thebibliography}{10}
\expandafter\ifx\csname url\endcsname\relax
  \def\url#1{\texttt{#1}}\fi
\expandafter\ifx\csname urlprefix\endcsname\relax\def\urlprefix{URL }\fi
\expandafter\ifx\csname href\endcsname\relax
  \def\href#1#2{#2} \def\path#1{#1}\fi

\bibitem{wolpaw2002bcireview}
J.~R. Wolpaw, N.~Birbaumer, D.~J. McFarland, G.~Pfurtscheller, T.~M. Vaughan,
  {Brain-computer interfaces for communication and control}, Clinical
  Neurophysiology 113~(6) (2002) 767--791.

\bibitem{PFU10}
G.~Pfurtscheller, B.~Z. Allison, C.~Brunner, G.~Bauernfeind,
  T.~Solis-Escalante, R.~Scherer, T.~O. Zander, G.~Mueller-Putz, C.~Neuper,
  N.~Birbaumer, The hybrid {BCI}, Frontiers in neuroscience 4.

\bibitem{ZAN11}
T.~O. Zander, C.~Kothe, Towards passive brain--computer interfaces: applying
  brain--computer interface technology to human--machine systems in general,
  Journal of Neural Engineering 8~(2) (2011) 025005+.

\bibitem{VID73}
J.~J. Vidal, Toward direct brain-computer communication., Annual review of
  biophysics and bioengineering 2~(1) (1973) 157--180.

\bibitem{BAY00}
J.~D. Bayliss, D.~H. Ballard, Single trial p3 epoch recognition in a virtual
  environment, Neurocomputing 32-33 (2000) 637--642.

\bibitem{LOP09}
M.~A. Lopez, H.~Pomares, F.~Pelayo, J.~Urquiza, J.~Perez, Evidences of
  cognitive effects over auditory steady-state responses by means of artificial
  neural networks and its use in brain--computer interfaces, Neurocomputing
  72~(16-18) (2009) 3617--3623.

\bibitem{HUA11}
Y.~Huang, D.~Erdogmus, M.~Pavel, S.~Mathan, K.~E. Hild, A framework for rapid
  visual image search using single-trial brain evoked responses, Neurocomputing
  74~(12-13) (2011) 2041--2051.

\bibitem{JRA11}
N.~Jrad, M.~Congedo, R.~Phlypo, S.~Rousseau, R.~Flamary, F.~Yger,
  A.~Rakotomamonjy, {sw-SVM}: sensor weighting support vector machines for
  {EEG}-based brain--computer interfaces, Journal of Neural Engineering 8
  (2011) 056004+.

\bibitem{TU12}
W.~Tu, S.~Sun, A subject transfer framework for {EEG} classification,
  Neurocomputing 82 (2012) 109--116.

\bibitem{NIE04}
E.~Niedermeyer, F.~Lopes~da Silva, Electroencephalography: Basic Principles,
  Clinical Applications, and Related Fields, 5th Edition, Lippincott Williams
  \& Wilkins, 2004.

\bibitem{Blankertz2006}
B.~Blankertz, K.-R. Muller, D.~Krusienski, G.~Schalk, J.~Wolpaw, A.~Schlogl,
  G.~Pfurtscheller, J.~Millan, M.~Schroder, N.~Birbaumer, {The BCI competition
  III: validating alternative approaches to actual BCI problems}, Neural
  Systems and Rehabilitation Engineering, IEEE Transactions on 14~(2) (2006)
  153--159.

\bibitem{BLA04}
B.~Blankertz, K.-R.~R. M\"{u}ller, G.~Curio, T.~M. Vaughan, G.~Schalk, J.~R.
  Wolpaw, A.~Schl\"{o}gl, C.~Neuper, G.~Pfurtscheller, T.~Hinterberger,
  M.~Schr\"{o}der, N.~Birbaumer, The {BCI} competition 2003: progress and
  perspectives in detection and discrimination of {EEG} single trials,
  Biomedical Engineering, IEEE Transactions on 51~(6) (2004) 1044--1051.

\bibitem{BLA06}
B.~Blankertz, K.~R. Muller, D.~J. Krusienski, G.~Schalk, J.~R. Wolpaw,
  A.~Schlogl, G.~Pfurtscheller, J.~Millan, M.~Schroder, N.~Birbaumer, The {BCI}
  competition {III}: validating alternative approaches to actual {BCI}
  problems, Neural Systems and Rehabilitation Engineering, IEEE Transactions on
  14~(2) (2006) 153--159.

\bibitem{TAN12}
M.~Tangermann, K.-R. M{\"u}ller, A.~Aertsen, N.~Birbaumer, C.~Braun,
  C.~Brunner, R.~Leeb, C.~Mehring, K.~J. Miller, G.~Mueller-Putz, G.~Nolte,
  G.~Pfurtscheller, H.~Preissl, G.~Schalk, A.~Schl{\"o}gl, C.~Vidaurre,
  S.~Waldert, B.~Blankertz, {Review of the BCI Competition IV}, Frontiers in
  Neuroscience 6~(55).

\bibitem{ALL10}
B.~Z. Allison, C.~Neuper, Could anyone use a {BCI}?, in: D.~S. Tan, A.~Nijholt
  (Eds.), Brain-Computer Interfaces, Human-Computer Interaction Series,
  Springer London, 2010, Ch.~3, pp. 35--54.

\bibitem{VID10}
C.~Vidaurre, B.~Blankertz, {Towards a cure for {BCI} illiteracy.}, Brain
  topography 23~(2) (2010) 194--198.

\bibitem{HAM12}
E.~M. Hammer, S.~Halder, B.~Blankertz, C.~Sannelli, T.~Dickhaus, S.~Kleih,
  K.-R. M{\"u}ller, A.~K{\"u}bler, {Psychological predictors of {SMR}-{BCI}
  performance}, Biological Psychology 89~(1) (2012) 80--86.

\bibitem{Obermaier2001}
B.~Obermaier, C.~Guger, C.~Neuper, G.~Pfurtscheller, {Hidden Markov models for
  online classification of single trial EEG data}, Pattern Recognition Letters
  22~(12) (2001) 1299--1309.

\bibitem{SHE06}
P.~Shenoy, M.~Krauledat, B.~Blankertz, R.~P.~N. Rao, K.-R. M\"{u}ller, Towards
  adaptive classification for {BCI}, Journal of Neural Engineering 3~(1) (2006)
  R13+.

\bibitem{Lenhardt2008}
A.~Lenhardt, M.~Kaper, H.~Ritter, {An Adaptive P300-Based Online Brain Computer
  Interface}, Neural Systems and Rehabilitation Engineering, IEEE Transactions
  on 16~(2) (2008) 121--130.

\bibitem{Bin2009AnOnlineMultiCh}
G.~Bin, X.~Gao, Z.~Yan, B.~Hong, S.~Gao, {An online multi-channel SSVEP-based
  brain-computer interface using a canonical correlation analysis method},
  Journal of Neural Engineering 6~(4).

\bibitem{ZHU11}
D.~Zhu, G.~Garcia-Molina, V.~Mihajlovi\'{c}, R.~Aarts, Online {BCI}
  implementation of {High-Frequency} phase modulated visual stimuli, in:
  C.~Stephanidis (Ed.), Universal Access in Human-Computer Interaction. Users
  Diversity, Vol. 6766 of Lecture Notes in Computer Science, Springer Berlin
  Heidelberg, 2011, pp. 645--654.

\bibitem{Lee2003}
H.~Lee, S.~Choi, {PCA+HMM+SVM for EEG pattern classification}, Signal
  Processing and Its Applications, 2003. Proceedings. Seventh International
  Symposium on 1 (2003) 541--544.

\bibitem{kalunga2013ssvep}
E.~Kalunga, K.~Djouani, Y.~Hamam, S.~Chevallier, E.~Monacelli, {SSVEP
  enhancement based on Canonical Correlation Analysis to improve BCI
  performances}, in: AFRICON, 2013, IEEE, 2013, pp. 1--5.

\bibitem{Lu2010Regular}
H.~Lu, H.-L. Eng, C.~Guan, K.~Plataniotis, A.~Venetsanopoulos, {Regularized
  Common Spatial Pattern With Aggregation for EEG Classification in
  Small-Sample Setting}, Biomedical Engineering, IEEE Transactions on 57~(12)
  (2010) 2936--2946.

\bibitem{BLA08b}
B.~Blankertz, R.~Tomioka, S.~Lemm, M.~Kawanabe, K.~R. Muller, {Optimizing
  Spatial filters for Robust {EEG} {Single-Trial} Analysis}, Signal Processing
  Magazine, IEEE 25~(1) (2008) 41--56.

\bibitem{LOT11}
F.~Lotte, C.~Guan, {Regularizing Common Spatial Patterns to Improve {BCI}
  Designs: Unified Theory and New Algorithms}, Biomedical Engineering, IEEE
  Transactions on 58~(2) (2011) 355--362.

\bibitem{ANG12}
K.~K.~K. Ang, Z.~Y.~Y. Chin, C.~Wang, C.~Guan, H.~Zhang, Filter bank common
  spatial pattern algorithm on {BCI} competition {IV} datasets 2a and 2b.,
  Frontiers in neuroscience 6.

\bibitem{YAN12a}
Y.~Yang, S.~Chevallier, J.~Wiart, I.~Bloch, {Automatic selection of the number
  of spatial filters for motor-imagery BCI}, in: M.~Verleysen (Ed.), {European
  Symposium on Artificial Neural Networks (ESANN)}, 2012, pp. 109--114.

\bibitem{Hardoon2004}
D.~R. Hardoon, S.~R. Szedmak, J.~R. Shawe-Taylor, {Canonical Correlation
  Analysis: An Overview with Application to Learning Methods}, Neural Comput.
  16~(12) (2004) 2639--2664.

\bibitem{Lin2007}
Z.~Lin, C.~Zhang, W.~Wu, X.~Gao, {Frequency Recognition Based on Canonical
  Correlation Analysis for SSVEP-Based BCIs}, Biomedical Engineering, IEEE
  Transactions on 53~(12) (2006) 2610--2614.

\bibitem{absil2009optimization}
P.-A. Absil, R.~Mahony, R.~Sepulchre, Optimization algorithms on matrix
  manifolds, Princeton University Press, 2009.

\bibitem{barachant2010riemannian}
A.~Barachant, S.~Bonnet, M.~Congedo, C.~Jutten, {R}iemannian geometry applied
  to {BCI} classification, in: Latent Variable Analysis and Signal Separation,
  Springer, 2010, pp. 629--636.

\bibitem{barachant2012multiclass}
A.~Barachant, S.~Bonnet, M.~Congedo, C.~Jutten, {Multiclass brain--computer
  interface classification by Riemannian geometry}, Biomedical Engineering,
  IEEE Transactions on 59~(4) (2012) 920--928.

\bibitem{panicker2010adaptation}
R.~C. Panicker, S.~Puthusserypady, Y.~Sun, Adaptation in p300 brain--computer
  interfaces: A two-classifier cotraining approach, Biomedical Engineering,
  IEEE Transactions on 57~(12) (2010) 2927--2935.

\bibitem{schettini2014self}
F.~Schettini, F.~Aloise, P.~Aric{\`o}, S.~Salinari, D.~Mattia, F.~Cincotti,
  Self-calibration algorithm in an asynchronous p300-based brain--computer
  interface, Journal of neural engineering 11~(3) (2014) 035004.

\bibitem{verschore2012dynamic}
H.~Verschore, P.-J. Kindermans, D.~Verstraeten, B.~Schrauwen, {Dynamic stopping
  improves the speed and accuracy of a P300 speller}, in: Artificial Neural
  Networks and Machine Learning--ICANN 2012, Springer, 2012, pp. 661--668.

\bibitem{barachant2012bci}
A.~Barachant, S.~Bonnet, M.~Congedo, C.~Jutten, et~al., {BCI} signal
  classification using a riemannian-based kernel, in: Proceeding of the 20th
  European Symposium on Artificial Neural Networks, Computational Intelligence
  and Machine Learning, 2012, pp. 97--102.

\bibitem{BAR13}
A.~Barachant, S.~Bonnet, M.~Congedo, C.~Jutten, Classification of covariance
  matrices using a riemannian-based kernel for {BCI} applications,
  Neurocomputing 112 (2013) 172--178.

\bibitem{yger2013review}
F.~Yger, {A review of kernels on covariance matrices for BCI applications}, in:
  Machine Learning for Signal Processing (MLSP), 2013 IEEE International
  Workshop on, IEEE, 2013, pp. 1--6.

\bibitem{jayasumana2013kernel}
S.~Jayasumana, R.~Hartley, M.~Salzmann, H.~Li, M.~Harandi, Kernel methods on
  the riemannian manifold of symmetric positive definite matrices, in: Computer
  Vision and Pattern Recognition (CVPR), 2013 IEEE Conference on, IEEE, 2013,
  pp. 73--80.

\bibitem{xie2013nonlinear}
Y.~Xie, J.~Ho, B.~Vemuri, On a nonlinear generalization of sparse coding and
  dictionary learning, in: Proceedings of the 30th International Conference on
  Machine Learning, NIH Public Access, 2013, p. 1480.

\bibitem{goh2008unsupervised}
A.~Goh, R.~Vidal, {Unsupervised Riemannian clustering of probability density
  functions}, in: {Machine Learning and Knowledge Discovery in Databases},
  Springer, 2008, pp. 377--392.

\bibitem{goh2008clustering}
A.~Goh, R.~Vidal, {Clustering and dimensionality reduction on Riemannian
  manifolds}, in: {Computer Vision and Pattern Recognition, 2008. CVPR 2008.
  IEEE Conference on}, IEEE, 2008, pp. 1--7.

\bibitem{PEN06}
X.~Pennec, P.~Fillard, N.~Ayache, A riemannian framework for tensor computing,
  International Journal of Computer Vision 66~(1) (2006) 41--66.

\bibitem{barachant2013riemannian}
A.~Barachant, A.~Andreev, M.~Congedo, et~al., The {R}iemannian potato: an
  automatic and adaptive artifact detection method for online experiments using
  {R}iemannian geometry, in: Proceedings of TOBI Workshop IV, 2013, pp. 19--20.

\bibitem{amari2010information}
S.-I. Amari, Information geometry in optimization, machine learning and
  statistical inference, Frontiers of Electrical and Electronic Engineering in
  China 5~(3) (2010) 241--260.

\bibitem{samek2013robust}
W.~Samek, D.~Blythe, K.-R. M{\"u}ller, M.~Kawanabe, Robust spatial filtering
  with beta divergence, in: Advances in Neural Information Processing Systems,
  2013, pp. 1007--1015.

\bibitem{samek2014information}
W.~Samek, K.-R. Muller, Information geometry meets bci spatial filtering using
  divergences, in: Brain-Computer Interface (BCI), 2014 International Winter
  Workshop on, IEEE, 2014, pp. 1--4.

\bibitem{barachant2011channel}
A.~Barachant, S.~Bonnet, Channel selection procedure using {R}iemannian
  distance for {BCI} applications, in: Neural Engineering (NER), 2011 5th
  International IEEE/EMBS Conference on, IEEE, 2011, pp. 348--351.

\bibitem{barachant2010common}
A.~Barachant, S.~Bonnet, M.~Congedo, C.~Jutten, Common spatial pattern
  revisited by {R}iemannian geometry, in: Multimedia Signal Processing (MMSP),
  2010 IEEE International Workshop on, IEEE, 2010, pp. 472--476.

\bibitem{congedo2013new}
M.~Congedo, A.~Barachant, A.~Andreev, A new generation of brain-computer
  interface based on riemannian geometry, arXiv preprint arXiv:1310.8115.

\bibitem{li2009eeg}
Y.~Li, K.~M. Wong, H.~De~Bruin, Eeg signal classification based on a riemannian
  distance measure, in: Science and Technology for Humanity (TIC-STH), 2009
  IEEE Toronto International Conference, IEEE, 2009, pp. 268--273.

\bibitem{li2012electroencephalogram}
Y.~Li, K.~Wong, H.~De~Bruin, Electroencephalogram signals classification for
  sleepstate decision-a riemannian geometry approach, Signal Processing, IET
  6~(4) (2012) 288--299.

\bibitem{lee2010introduction}
J.~Lee, Introduction to topological manifolds, Vol. 940, Springer, 2010.

\bibitem{jost2011riemannian}
J.~Jost, Riemannian geometry and geometric analysis, Vol. 62011, Springer,
  2011.

\bibitem{moakher2005differential}
M.~Moakher, A differential geometric approach to the geometric mean of
  symmetric positive-definite matrices, SIAM Journal on Matrix Analysis and
  Applications 26~(3) (2005) 735--747.

\bibitem{fletcher2004principal}
P.~T. Fletcher, C.~Lu, S.~M. Pizer, S.~Joshi, Principal geodesic analysis for
  the study of nonlinear statistics of shape, Medical Imaging, IEEE
  Transactions on 23~(8) (2004) 995--1005.

\bibitem{fukunaga1990introduction}
K.~Fukunaga, Introduction to statistical pattern recognition, Academic press,
  1990.

\bibitem{ledoit2004well}
O.~Ledoit, M.~Wolf, A well-conditioned estimator for large-dimensional
  covariance matrices, Journal of multivariate analysis 88~(2) (2004) 365--411.

\bibitem{blankertz2011single}
B.~Blankertz, S.~Lemm, M.~Treder, S.~Haufe, K.-R. M{\"u}ller, Single-trial
  analysis and classification of erp components—a tutorial, NeuroImage 56~(2)
  (2011) 814--825.

\bibitem{schafer2005shrinkage}
J.~Sch{\"a}fer, K.~Strimmer, A shrinkage approach to large-scale covariance
  matrix estimation and implications for functional genomics, Statistical
  applications in genetics and molecular biology 4~(1).

\bibitem{pascal123theoretical}
F.~Pascal, P.~Forster, J.~P. Ovarlez, P.~Arzabal, Theoretical analysis of an
  improved covariance matrix estimator in non-gaussian noise, in: IEEE
  International Conference on Acoustics, Speech, and Signal Processing
  (ICASSP)., Vol.~4, 2005.

\bibitem{pencina2008evaluating}
M.~J. Pencina, R.~B. D'Agostino, R.~S. Vasan, Evaluating the added predictive
  ability of a new marker: from area under the roc curve to reclassification
  and beyond, Statistics in medicine 27~(2) (2008) 157--172.

\end{thebibliography}
\end{document}